\documentclass[letterpaper]{article} %
\input{versions} %
\ifarxiv
\usepackage[preprint]{aaai2027}
\else
\usepackage[submission]{aaai2027}  %
\fi
\usepackage[hyphens]{url}  %
\usepackage{graphicx} %
\usepackage{natbib}  %
\usepackage{caption} %
\usepackage{amsmath}
\usepackage{amssymb}

\usepackage{algorithm}
\usepackage{algorithmicx}
\usepackage{algpseudocode}
\usepackage{algcompatible}

\usepackage{booktabs}
\usepackage{multirow}
\usepackage{arydshln}
\usepackage{subcaption}

\usepackage{tcolorbox}
\tcbuselibrary{breakable} %
\tcbset{colback=gray!5,colframe=gray!80!black,boxrule=0.5mm,arc=4mm,breakable}

\usepackage{xspace}

\usepackage{cleveref}
\crefformat{section}{\S#2#1#3}
\crefformat{subsection}{\S#2#1#3}
\crefformat{subsubsection}{\S#2#1#3}
\Crefformat{section}{\S#2#1#3}
\Crefformat{subsection}{\S#2#1#3}
\Crefformat{subsubsection}{\S#2#1#3}

\usepackage{algorithm}
\usepackage{algpseudocode}
\usepackage{amsmath}

\ifdefined\revisionmode
\newcommand{\yiwei}[1]{{\color{brown} {\bf Yiwei:} #1}}
\newcommand{\jieyu}[1]{{\color{orange} {\bf Jieyu:} #1}}
\newcommand{\ziyun}[1]{{\color{red} {\bf Ziyun:} #1}}
\newcommand{\huapeng}[1]{{\color{magenta} {\bf Huapeng:} #1}}
\newcommand{\yi}[1]{{\color{purple} {\bf Yi:} #1}}
\newcommand{\phil}[1]{{\color{blue} {\bf Phil:} #1}}
\else
\newcommand{\yiwei}[1]{}
\newcommand{\ziyun}[1]{}
\newcommand{\jieyu}[1]{}
\newcommand{\huapeng}[1]{}
\newcommand{\yi}[1]{}
\newcommand{\phil}[1]{}
\fi

\newcommand{\ourname}{AdaDINO\xspace}

\newcommand{\VisionTower}{Vision Encoder\xspace}
\newcommand{\TextTower}{Text Encoder\xspace}

\newcommand{\defn}[1]{\textit{\textbf{#1}}}

\newcommand{\myparagraph}[1]{\smallskip\noindent {\bf #1.}}

\newcommand{\hide}[1]{}

\newcommand{\RNum}[1]{\expandafter{\romannumeral #1\relax}}
\newcommand\romenum[1]{\mbox{(\textit{\RNum{#1}})}\nolinebreak{}}

\newcommand{\TinyNAS}{\textsc{Tiny}\xspace}
\newcommand{\SmallNAS}{\textsc{Small}\xspace}
\newcommand{\MinNAS}{\textsc{Min}\xspace}
\newcommand{\BaseNAS}{\textsc{Base}\xspace}
\newcommand{\LargeNAS}{\textsc{Large}\xspace}

\ifsplitparts
\let\cite\citep
\usepackage{bibunits}
\defaultbibliographystyle{aaai2027}
\defaultbibliography{ref}
\fi

\newcommand{\realTitle}{\ourname: Context-Adaptive DINO-Distilled Vision Foundation Models for Efficient Open-Vocabulary Edge Inference}

\ifshowmain
\title{\realTitle}
\else
\title{\realTitle\xspace(Supplementary Material)}
\fi

\author{
    Yiwei Zhao\textsuperscript{\rm 1},
    Yi Zheng\textsuperscript{\rm 2},
    Huapeng Su\textsuperscript{\rm 2},
    Jieyu Lin\textsuperscript{\rm 2},
    Stefano Ambrogio\textsuperscript{\rm 2},
    Cijo Jose\textsuperscript{\rm 2},\\
    Micha\"el Ramamonjisoa\textsuperscript{\rm 2},
    Patrick Labatut\textsuperscript{\rm 2},
    Barbara De Salvo\textsuperscript{\rm 2},
    Chiao Liu\textsuperscript{\rm 2},\\
    Phillip B. Gibbons\textsuperscript{\rm 1},
    Ziyun Li\textsuperscript{\rm 2}
}
\affiliations{
    \textsuperscript{\rm 1}Carnegie Mellon University\quad{}\textsuperscript{\rm 2}Meta
}

\begin{document}

\ifarxiv
\maketitle
\begin{abstract}
Always-on contextual AI runs language-aligned vision foundation models (VFMs) on edge devices, where the on-device model is the dominant continuous compute cost under strict latency and power limits.
Due to an observed low-frequency shift in scene context and its relevant vocabulary, we present \ourname, an adaptive framework that makes on-device VFM inference efficient by matching execution to the current scene and task.
We build on a known phenomenon, that the accuracy drop of shrinking model sizes depends on the task, and turn it into task-level adaptive execution.
\ourname integrates neural architecture search (NAS) into a language-aligned VFM backbone distilled from DINOv2, training a single family of subnets for efficient execution during runtime.
A multimodal large language model (LLM) on the cloud, invoked at low frequency, refines the candidate class set from scene context, while a learned selector activates the least-cost subnet predicted to retain a target fraction of accuracy.
With the backbone and semantic pipeline held fixed, learned selection alone reduces average compute by $37\%$ over the best fixed subnet at equal segmentation accuracy.
Across zero-shot classification and open-vocabulary segmentation, \ourname establishes a strong accuracy-efficiency frontier, improving over evaluated models of comparable sizes by up to $7.9\%$ in acc@1 on IN1K and $5.2\%$ mIoU on ADE20K, and reducing average FLOPs by up to $74.9\%$ at similar accuracy.
\end{abstract}
\section{Introduction}
\label{sec:intro}

Advances in smartphones, wearable devices~\cite{abbaspourazad2024largescale}, augmented and virtual reality~\cite{abrash2021creating}, robotics, and autonomous driving are pushing multimodal AI onto edge devices, driving interest in \textit{always-on contextual AI}---continuous, context-aware systems running persistently under tight energy and latency constraints~\cite{wang2024lifelong}.
Such systems already ship as real products.
Smart glasses like Meta Ray-Ban~\cite{meta_rayban_smart_glasses} enable contextual AI by forwarding multimodal queries to the cloud, costing multi-second responses and short battery life.

Industry (such as Meta) is converging on an edge-cloud split as in Fig.\ref{fig:intro_motivation}: the edge runs a lightweight \textbf{vision foundation model} (VFM) for real-time perception, while heavy or knowledge-intensive tasks go to a cloud multimodal LLM (MM-LLM)~\cite{Chen2023blip2}.
The split works because context changes far more slowly than frames arrive: egocentric scenes and activities typically persist for minutes or longer~\cite{grauman2022ego4d,grauman2024egoexo4d}.

\begin{figure}[t]
    \centering
    \includegraphics[width=0.65\linewidth]{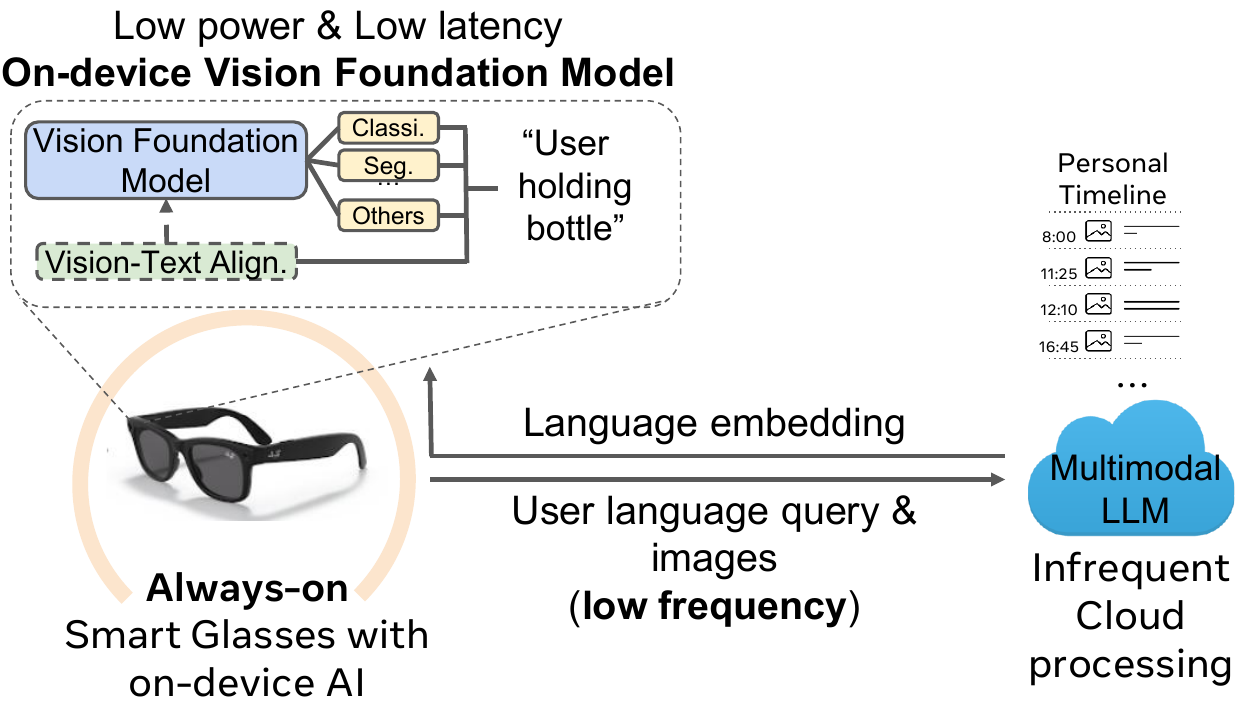}
    \vspace{-0.3em}
    \caption{Motivating deployment scenario: always-on smart glasses with an on-device VFM.}
    \label{fig:intro_motivation}
    \vspace{-0.5em}
\end{figure}

Building this stack end to end is an industry-scale effort. We focus on one important layer of this stack that stands on its own as a machine-learning problem: the dominant and continuous compute cost of the always-on on-device vision model.
Language-aligned VFMs are large and computation-intensive, but edge devices operate under strict compute, memory, and power limits with interaction budgets of 200--1000~ms~\cite{stein2021latency}.
Direct VFM execution under such constraints is infeasible, and prior attempts that simply shrink model capacity~\cite{wu2023tinyclip} often suffer large performance drops (see \S\ref{subsec:bg_clip}).

\begin{figure*}[t]
    \centering
    \vspace{-1em}
    \begin{minipage}[b]{0.71\textwidth}
        \centering
        \includegraphics[width=\linewidth]{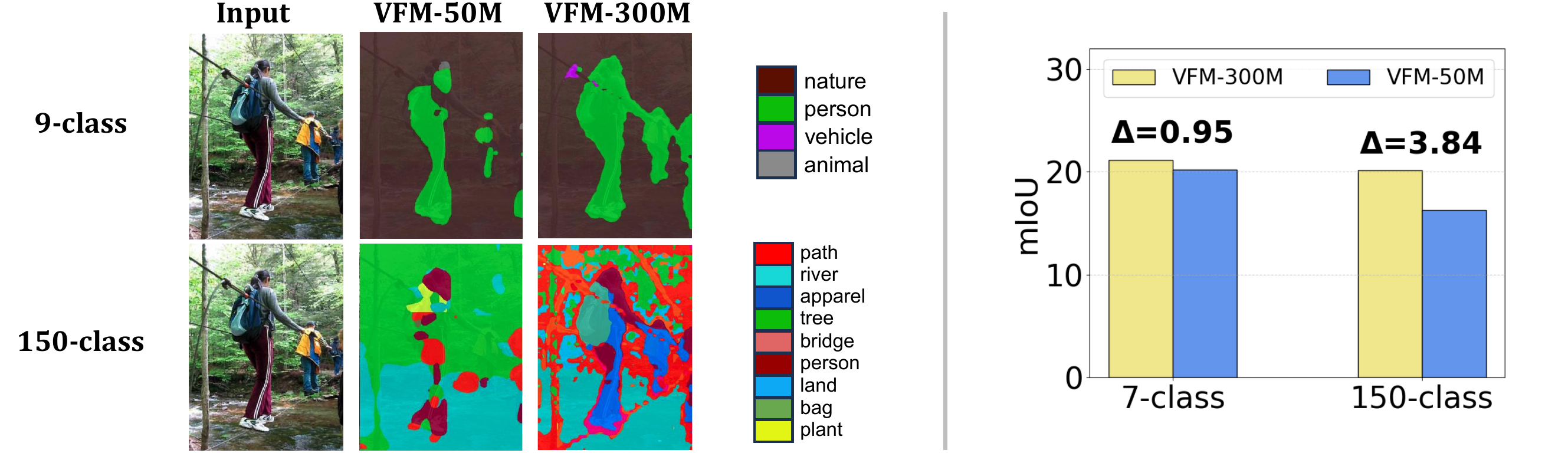}
    \end{minipage}\hfill
    \begin{minipage}[b]{0.29\textwidth}
        \centering
        \includegraphics[width=0.93\linewidth]{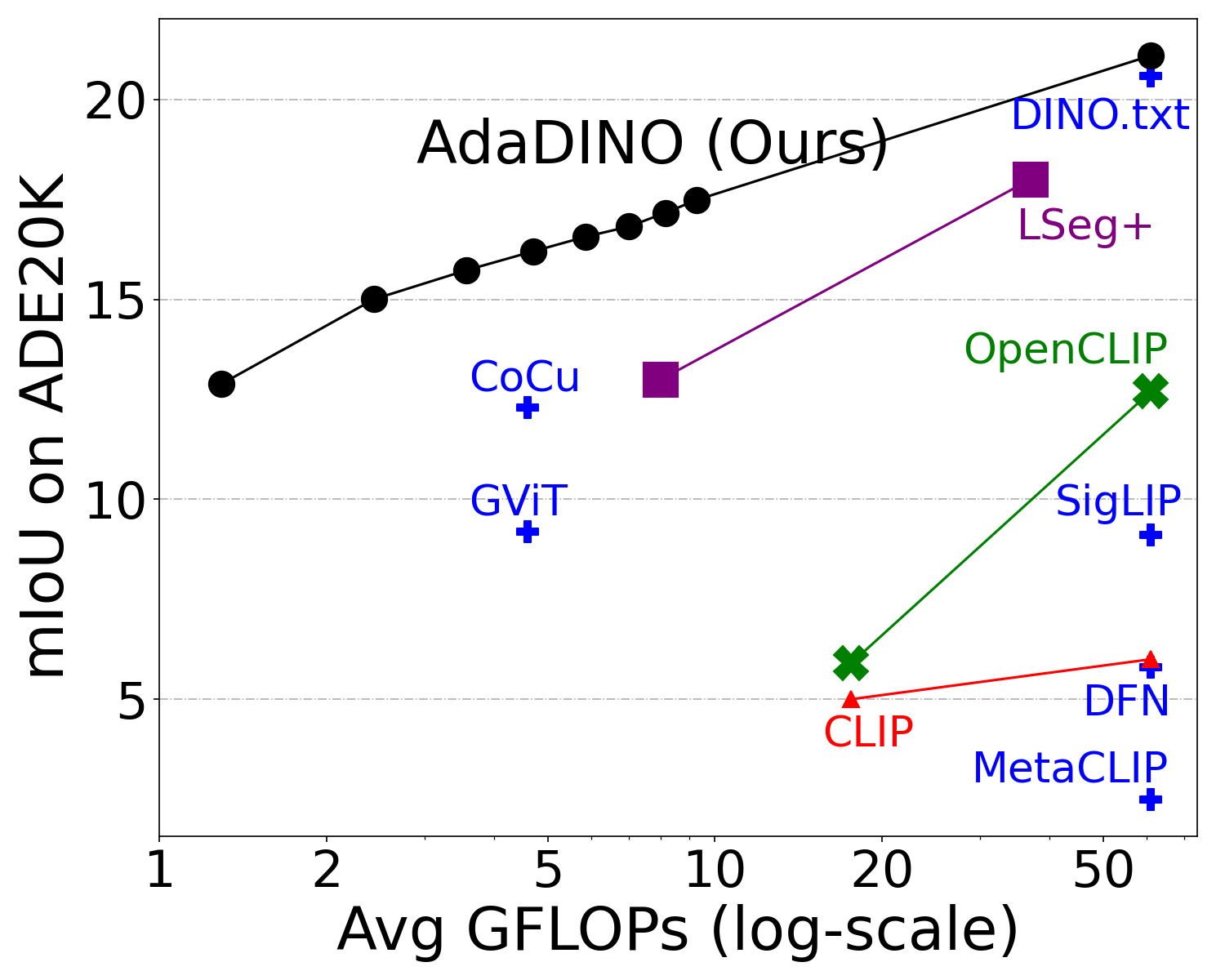}
    \end{minipage}\\
    \begin{minipage}[t]{0.71\textwidth}
        \caption{Open-vocabulary segmentation on ADE20K~\cite{zhou2019semantic,zhou2017scene} using 50M and 300M-parameter VFMs (300M: DINO.txt; 50M: similarly distilled and fine-tuned). Results are shown for both the original 150-class and the grouped 9-class setting (grouping based on WordNet~\cite{fellbaum1998wordnet}, described in \suppref{sec:supp_granularity}).
    \textbf{Left}: Both models perform similarly on the simpler 9-class task, but in the complex 150-class setting, VFM-300M yields \textit{clear object boundaries} while VFM-50M gives \textit{blurred} ones.
    \textbf{Right}: Going from 300M to 50M costs only 0.95 mIoU on the 9-class task but 3.84 on the 150-class task.}
    \label{fig:intro_granularity}
    \end{minipage}\hfill
    \begin{minipage}[t]{0.26\textwidth}
        \caption{End-to-end open-vocabulary ADE20K segmentation. \ourname{} improves mIoU by up to $5.2\%$ and cuts FLOPs by up to $74.9\%$ over prior models.}
        \label{fig:intro_main_result}
    \end{minipage}
    \vspace{-1em}
\end{figure*}

However, not all tasks need a full VFM---simpler tasks suit smaller models.
As in \suppref{tab:intro_dino}, DINOv2 degrades little on simpler datasets such as Pets~\cite{parkhi2012pets} and Cal101~\cite{li2008cal101} with smaller models (drops of at most 1.6), while harder datasets incur larger drops (up to 5.6).

The same pattern holds in open-vocabulary segmentation: in Fig.~\ref{fig:intro_granularity}, 50M and 300M VFMs are comparable on a grouped 9-class ADE20K task but diverge significantly on the full 150-class task (details in \suppref{sec:supp_granularity}).
This motivates \textbf{\textit{adaptive model selection}} on the edge based on \textbf{\textit{task difficulty}}, rather than a single fixed configuration.

In this paper, we study a representative VFM family---DINOv2~\cite{oquab2024dinov2} and the CLIP-style DINO.txt~\cite{jose2025dinov2} framework---and \textit{adapt it for the edge}.
We leverage \textbf{\textit{the sensitivity of open-vocabulary tasks to model capacity}} for efficient edge deployment of DINO-based VFMs.
A fixed edge VFM overkills: it spends full capacity on easy vocabularies and compares against a broad class set even when context rules out most classes.
We introduce \textbf{\textit{task-level adaptive selection}}, which chooses the execution scheme fitting the current task and resources, and integrate Neural Architecture Search (NAS) into a two-stage pipeline of DINOv2 distillation and DINO.txt-style vision-text alignment.
In principle the pipeline generalizes to any distillable VFM with a CLIP-style framework, though we instantiate it on DINO as a representative, well-performing choice.

To improve edge model efficiency, an LLM-powered runtime manager, invoked at low frequency, infers scene context, filters out implausible classes, and supplies an execution scheme under current scene or vocabulary.
\ourname{} targets one layer in the full system stack---making the on-device VFM efficient using this runtime manager.
When to refresh it, and how the full device behaves over long deployments, belong to the surrounding stack and lie outside the ML scope.

We present \textbf{\ourname}, an adaptive edge VFM with a training pipeline and a proof-of-concept deployment on real edge hardware.
It brings DINO-family VFMs to the edge via task-context adaptation on standard deployable components.
Our contributions are:
\romenum{1} NAS-integrated adaptive VFM deployment for efficient edge execution;
\romenum{2} LLM-based semantic understanding for refined class sets and text embeddings; and
\romenum{3} learned task-level subnet selection, which alone cuts average compute by more than a third at matched accuracy.
Evaluations (Fig.\ref{fig:intro_main_result}, Tab.\ref{tab:eval_zeroshot}) show that \ourname achieves a better accuracy-efficiency trade-off than prior work~\cite{ghiasi2022scaling,jose2025dinov2,radford2021CLIP,fang2024data,ilharco2021openclip,xing2023rewrite,xu2024demystifying,xu2023learning,zhai2023sigmoid,roth2023waffling,mirza2023lafter,khattak2025learning}.

\section{Background and Related Work}
\label{sec:bg}
This section summarizes the work most relevant to \ourname{}.
An extended discussion of related work in each topic is provided in \suppref{sec:supp_related}.

\subsection{Always-on Contextual AI on Edge Devices}
\label{subsec:bg_contextual_ai}
Wearable edge platforms are motivating \emph{always-on contextual AI} that operates continuously under strict power, thermal, and memory constraints.
Persistent offloading is impractical at wearable hardware budgets~\cite{lane2015deepx}, so a lightweight \emph{on-device vision foundation model} handles real-time perception such as classification and segmentation, while multimodal reasoning and broad world knowledge are delegated to a cloud-based multimodal LLM (Fig.\ref{fig:intro_motivation}).
Building such a device end-to-end---silicon, optics, batteries, interaction---is a division-scale industry effort, and deployed stacks already isolate the always-on perception model as a distinct layer from the intermittent cloud reasoning around it. We take that industry partition as given (\S\ref{sec:intro}, \suppref{subsec:supp_scope}) and address the layer it exposes: making the dominant always-on compute cost of on-device vision models spend only what the current task requires.

\subsection{Vision Foundation Models}
\label{subsec:bg_vfm}
Vision foundation models offer unified representations that generalize across tasks and domains, typically trained at scale with self-supervised learning (SSL).
Recent advances such as DINOv2~\cite{oquab2024dinov2} and iBOT~\cite{Zhou2022ibot} learn transferable features that rival or surpass those obtained via supervised pretraining methods like CLIP.
Large-scale VFMs, however, incur substantial compute, memory, and energy costs.
We present \emph{adaptive} VFMs that bring strong accuracy-efficiency trade-offs to the edge devices.

\subsection{Contrastive Learning for VFMs}
\label{subsec:bg_clip}

Contrastive learning methods such as CLIP~\cite{radford2021CLIP} enable zero-shot and open-vocabulary capabilities by aligning visual and textual representations.
A standard CLIP architecture consists of a \VisionTower and a \TextTower, both too costly for resource-constrained edge devices. Pruning or shrinking them often causes large accuracy loss.

A key property of CLIP for edge deployment is the \defn{differing frequencies} of \VisionTower{} and \TextTower{} calls.
In always-on contextual AI (Fig.~\ref{fig:intro_motivation}), prompts or scenes change infrequently, while perception runs continuously:

\begin{itemize}
    \item \textbf{\VisionTower (high-frequency):} Processes every incoming frame and must meet strict low-latency requirements (typically $200$--$1000$~ms).
    \item \textbf{\TextTower (low-frequency):} Invoked only when user prompts or scenes change, tolerating much higher latency.
\end{itemize}

This \textit{frequency asymmetry} creates a design opportunity: the \VisionTower must be heavily optimized for real-time edge inference, while the \TextTower can remain larger without impacting latency.

\myparagraph{Efficient VFMs and CLIP}
Prior work lowers VFM cost either \emph{statically}---distilling or shrinking CLIP-style encoders into compact backbones~\cite{vasu2024mobileclip}---or through a \emph{different notion of adaptiveness} than ours, in which a single fixed model spends more or less compute according to per-input difficulty, e.g., via token pruning and merging~\cite{liang2022evit, bolya2023tome, wu2025patchranking} or token/block halting~\cite{yin2022avit}.
All of these react to per-sample \emph{confidence} within a \emph{single, fixed, closed-set} model.
Our \ourname{} adapts along an orthogonal axis: it switches among \emph{open-vocabulary} subnets conditioned on \emph{scene- and task-level context} rather than per-frame confidence, and stays composable with the above, which can still operate within whichever subnet it activates.

\subsection{Neural Architecture Search}
NAS provides a principled framework for adaptive-capacity models suited to edge deployment.
Its key merit is support for \textbf{\textit{runtime subnet selection}} over sub-architectures of varying computational cost: at inference, a subnet is selected by task complexity or resources, enabling flexible accuracy-efficiency trade-offs.
Weight-sharing supernets~\cite{cai2020once} and runtime width switching in slimmable networks~\cite{yu2019slimmable,li2021dsnet} made this practical.
Rather than a new supernet-training algorithm, our contribution is an OFA-style supernet made compatible with foundation-model distillation and vision-text alignment, driven by scene-conditioned vocabulary refinement and text-set-conditioned subnet selection.

\section{Method}
\label{sec:method}

\subsection{Overall System Design}
\label{subsec:method_system}
We show the \ourname{} runtime system in Fig.\ref{fig:method_system}.
Our optimization target is the on-device vision model: given a low-frequency context signal, make it spend only what the current task needs. The design has two complementary components: 
(1) an \textbf{adaptive vision foundation model} deployed on the \textit{edge} device, running always-on for high-frequency real-time vision inference; and 
(2) a \textbf{cloud-side runtime manager}, invoked infrequently, where a multimodal LLM provides scene and context understanding and text embeddings, and a learned selector picks the edge subnet.
Together they adapt open-vocabulary inference as scenes and tasks change.
\ourname{} operates at the \emph{task-context} granularity: each context update defines a candidate class set that stays active across many subsequent frames. The context signal is supplied at low frequency, and deciding when to refresh it is an orthogonal deployment problem outside our ML scope.

\begin{figure*}
    \centering
    \includegraphics[width=0.56\textwidth]{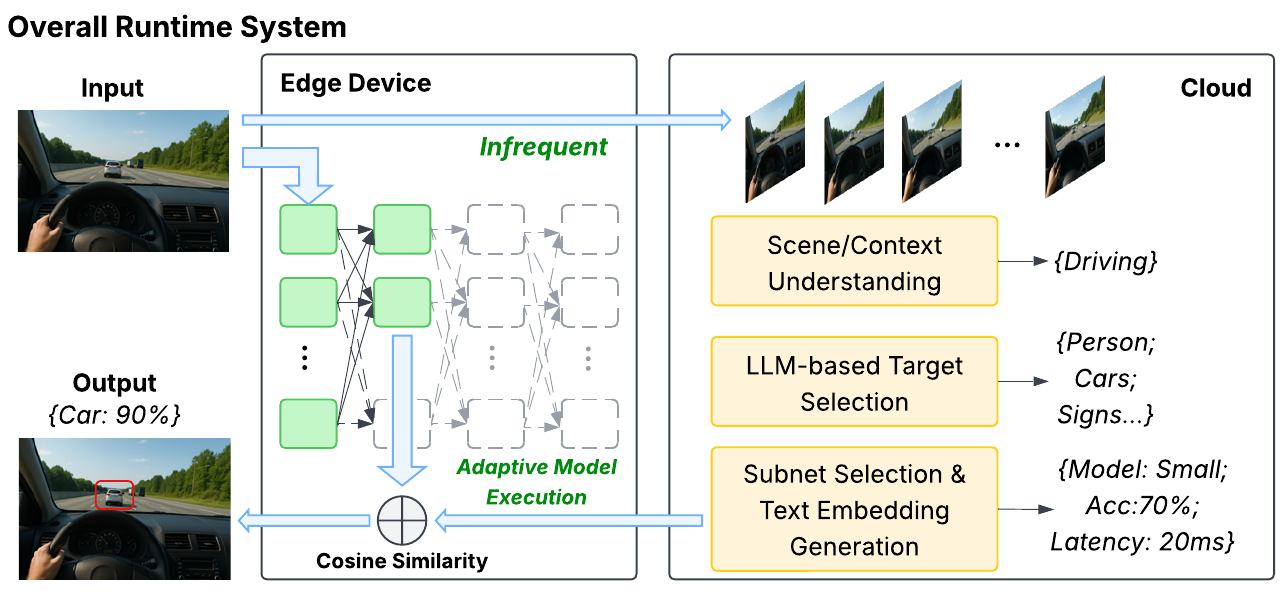}
    \vspace{-0.7em}
    \caption{Overview of the proposed \ourname. \textbf{Left}: Edge-side execution, where the adaptive \VisionTower follows cloud-side manager instructions to select an efficient execution scheme and perform vision-text contrastive inference. \textbf{Right}: Cloud-side execution, where the manager uses scene/context information to generate semantic understanding for the \TextTower and execution guidance for the \VisionTower.}
    \vspace{-0.8em}
    \label{fig:method_system}
\end{figure*}

\subsection{On-Edge Adaptive VFM}
\label{subsec:method_adaptive_vfm}

\myparagraph{Execution Flow}
As shown in Fig.~\ref{fig:method_system}, the cloud-side multimodal LLM receives a temporally sparse stream of images and/or user text interactions to infer the \emph{scene and context} (e.g., \textit{driving}).
Given this context, the LLM identifies relevant open-vocabulary targets (e.g., \textit{person, cars, signs}) and, using augmented model metadata from Fig.~\ref{fig:method_manager_flow}, the manager produces two outputs:
\romenum{1} a set of \emph{text embeddings} for the selected concepts, and
\romenum{2} an \emph{execution scheme} computed by the learned selector of Step III in \S\ref{subsec:method_manager} from the refined class set, telling which edge subnet to use.
The edge-side \VisionTower operates continuously on every frame, computing cosine similarity between vision and cloud-provided text embeddings for real-time open-vocabulary inference.

\myparagraph{Selective Execution Scheme}
As illustrated on the left side of Fig.~\ref{fig:method_system}, \ourname{} enables dynamic adaptation on the edge through a NAS-based supernet space that exposes multiple execution schemes with different computational costs.
At runtime, the system selects a subnet from this space based on the prediction of the manager, which reflects current task difficulty and latency-accuracy requirements.
The chosen subnet stays active for the next few rounds of inferences.

\myparagraph{Open-Vocabulary Vision Tasks}
With the subnet selected and text embeddings supplied by the cloud, the \VisionTower performs open-vocabulary vision tasks using a CLIP-style pipeline, similar to LiT~\cite{zhai2022lit} and DINO.txt. Each incoming frame is encoded into visual tokens and compared against the text embeddings using cosine similarity. For zero-shot classification, the comparison is performed using the global [\textsc{CLS}] token, whereas for open-vocabulary dense tasks such as segmentation, patch-level embeddings are contrastively matched to produce per-pixel similarity maps.

\begin{figure*}[t]
\centering
\begin{subfigure}[t]{0.40\linewidth}
    \centering
    \includegraphics[width=0.86\linewidth]{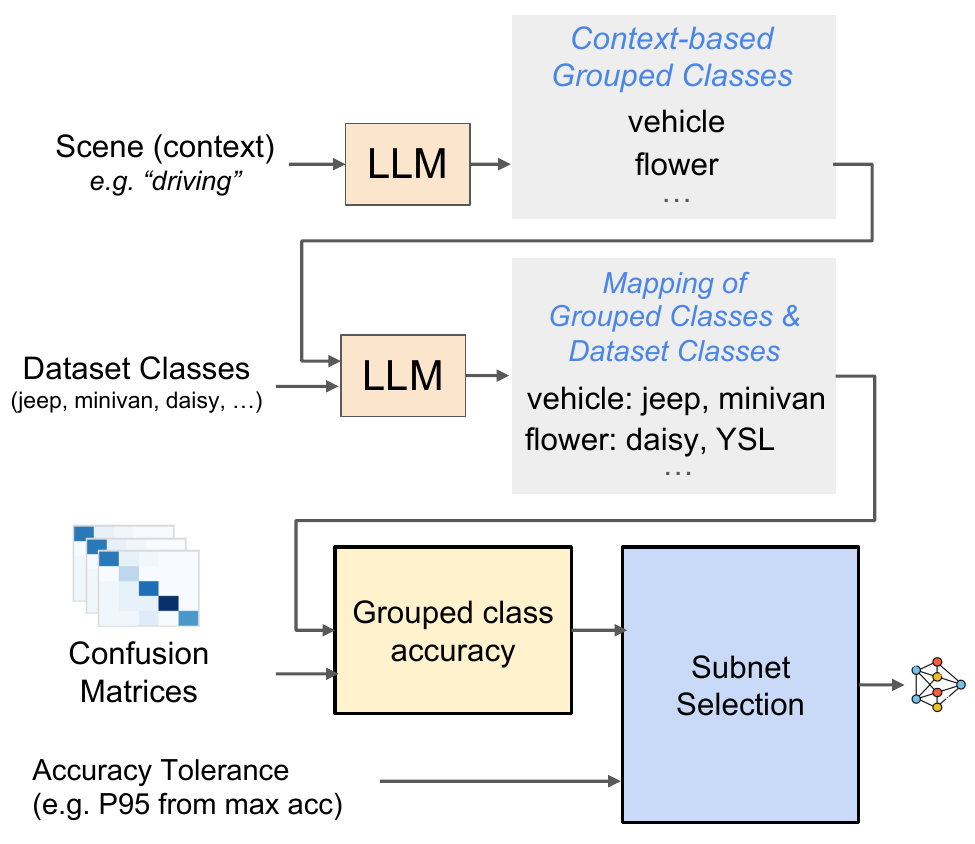}
    \phantomcaption
    \label{fig:method_manager_flow}
\end{subfigure}
\hfill
\begin{subfigure}[t]{0.52\linewidth}
    \centering
    \includegraphics[width=0.86\linewidth]{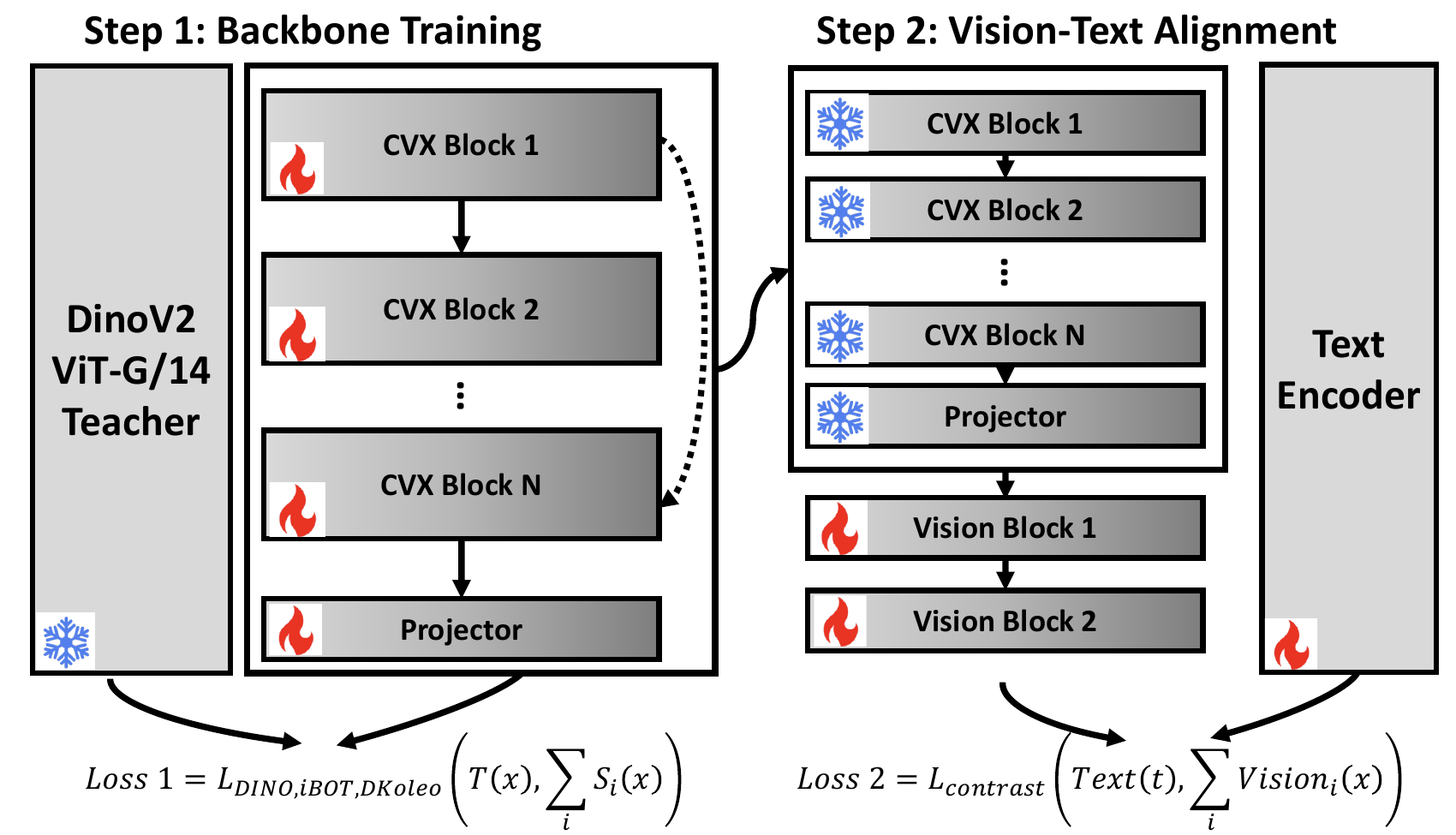}
    \phantomcaption
    \label{fig:train_nas_pipeline}
\end{subfigure}
\vspace{-1em}
\caption{
\textbf{Left (a)}: Operation flow of the LLM-assisted runtime manager (subnet choice follows Alg.~\ref{alg:selection}).
\textbf{Right (b)}: Overview of the training pipeline. The vision backbone is first distilled from a foundation model (DINOv2), followed by DINO.txt-style vision-text alignment. Both stages employ NAS and sandwich sampling~\cite{yu2019universally}.
}
\label{fig:method_pipeline}
\vspace{-0.8em}
\end{figure*}

\subsection{LLM-Guided Efficient Runtime Execution}
\label{subsec:method_manager}

A central component of \ourname{} is the \textbf{LLM-guided runtime manager}, invoked at low frequency (e.g., every few minutes).
As in Fig.\ref{fig:method_manager_flow}, this manager is responsible for:
\romenum{1} giving a \textit{semantic understanding} of the current scene to guide the \TextTower, and 
\romenum{2} selecting \textit{efficient} execution schemes of the \VisionTower{} via a learned selector (Step III).

From the low-frequency contextual signals described above, the LLM infers high-level scene context (e.g., \textit{``driving''}) and generates a set of \emph{context-based grouped classes}, as illustrated in the top block of Fig.~\ref{fig:method_manager_flow}. These coarse concepts are then mapped to fine-grained categories guided by vision datasets (e.g., \textit{vehicle} $\rightarrow$ \{\textit{jeep, minivan}\}) using a second LLM query, yielding open-vocabulary scene-related targets.

\myparagraph{Step I: Scene Annotation Generation}
For controlled and reproducible evaluation, we treat each image in public benchmarks as one context instance: the image is fed to a cloud-side MM-LLM, which produces a concise scene annotation within three words (e.g., \textit{``airport''} or \textit{``office''}). See \suppref{subsec:supp_annotation_prompts} for implementation details.
This protocol isolates the effect of semantic context on model execution, validated one context at a time.
In deployment, one annotation is reused across temporally adjacent frames at low frequency (e.g., every few minutes) shown in Fig.\ref{fig:method_manager_flow}, a schedule the projections below assume rather than measure.

\myparagraph{Step II: Semantic Understanding of a Scene}
Zero-shot classification and open-vocabulary segmentation in CLIP-style models require passing candidate class names into the \TextTower.
Instead of blindly enumerating the entire vocabulary, our runtime manager uses LLM-driven semantic reasoning to filter out implausible classes based on the task context, reducing noise and resolving ambiguous class definitions.
For example, given a scene context \textit{``sports field''}, the manager selects \textit{``field''} rather than \textit{``grass''} as the more contextually appropriate phrase.
This filtering step corresponds to the first stage of Fig.\ref{fig:method_manager_flow}, where dataset classes are aligned to grouped semantic categories.

\myparagraph{Step III: Difficulty-Based Model Selection}
The size and composition of the class set reflect the \emph{difficulty} of an open-vocabulary task, which can guide the runtime selection of an appropriately sized model for efficient edge execution.

A selector \emph{learns} this ability in an open-vocabulary way from the candidate class names, following CLIP-style methods that use text embeddings as open-vocabulary classifiers.
Specifically, the class names $C$ are embedded by the Text Encoder and summarized as $d=\textsc{SetStats}(\textsc{TextEnc}(C))$, an order-/size-independent statistic of their pairwise similarities, inspired by permutation-invariant set representations~\cite{zaheer2017deep}.
A capacity-monotone gradient-boosted regressor then predicts the \emph{accuracy-retention ratio} $\hat{r}_M\!\approx\!\mathrm{Acc}(M,C)/\mathrm{MaxAcc}$ of each subnet $M$, following prior learning-to-rank/budgeted-prediction formulations~\cite{li2007mcrank,xu2012greedymiser}.

Alg.~\ref{alg:selection} returns the least-cost subnet with cost $c(M)$ predicted to satisfy $\hat{r}_M\ge\alpha$, which can generalize to unseen class sets.
The selector is agnostic to where the class set comes from: an LLM is used in our prototype because it handles open-ended scene descriptions naturally, but other context providers can drive the same selector if necessary.

\begin{algorithm}[t]
\caption{Learned Scene-Aware Subnet Selection}
\label{alg:selection}
\begin{algorithmic}[1]
\REQUIRE class names $C$; tolerance $\alpha$; subnet $M$ \& cost $c(\cdot)$
\STATE $d \gets \textsc{SetStats}(\textsc{TextEnc}(C))$
\STATE $\hat{r}_M \gets \textsc{Retention}(d, M) \quad \forall M$
\STATE \textbf{return} $\arg\min\{\, c(M) : \hat{r}_M \ge \alpha \,\}$ \textbf{else} $M_{\max}$
\end{algorithmic}
\end{algorithm}

\myparagraph{Conceptually Reducing System Overhead}
Whereas prior contextual AI systems forward every multimodal query to the cloud (\S\ref{sec:intro}), our system invokes cloud MM-LLM at a much lower frequency (a few minutes).
As a conceptual example, the cloud-side call has a 6.4s P95 latency on our prototype hardware (\suppref{sec:supp_hardware}), small relative to the refresh interval.
Only vision tasks run in real time on-device.
This design incurs sustainable cost on the efficient edge subnet.
The resulting battery-life and cloud-cost gains are first-order projections under this cadence rather than measured results, reported as deployment motivation in \suppref{subsec:supp_scope}.

\myparagraph{System Robustness}
Cloud unavailability does not break \ourname{}: the edge VFM falls back to the largest subnets, which remain more efficient than prior baselines (see \S\ref{sec:eval}).

\section{Training}
\label{sec:train}

We integrate NAS into the DINO-based VFM, as illustrated in Fig.\ref{fig:train_nas_pipeline}, following the motivation of \S\ref{sec:intro}---a trend widely observed in prior work~\cite{lu2021neural} and verified in our open-vocabulary setting (\suppref{sec:supp_granularity}).
Our goal is to train a single \emph{supernet}, with subnets of varying computational costs selected at runtime.

\myparagraph{Stage 1: Backbone Distillation}  
We use OFA-NAS~\cite{cai2020once}, training a supernet over a wide range of widths and depths.
As in Fig.\ref{fig:train_nas_pipeline} (left), we distill a DINOv2 ViT-G/14 teacher into a ConvNeXt-based~\cite{Liu2022convnext, woo2023convnextv2} supernet.
ConvNeXt is used for edge deployability: its convolutional operators map onto fixed-function NPUs far better than the ViT teacher's attention (\S\ref{subsec:eval_sota}, \suppref{sec:supp_hardware}), and its structure keeps NAS integration much more tractable than transformer search spaces~\cite{serianni2023training,li2021bossnas}.

To ensure consistent output dimensionality across all subnets, we add a lightweight linear projector on top of the ConvNeXt backbone.
During this stage, all ConvNeXt blocks and the projector are trainable (indicated by flame icons in the figure). We optimize a DINO-style distillation loss $\mathcal{L}_1 = \sum_i \mathcal{L}_{\text{DINO},\text{iBOT},\text{Koleo}}\big(T(x), S_i(x)\big)$, where $T(\cdot)$ is teacher features and $S_i(\cdot)$ is outputs of sampled subnets.

\myparagraph{Stage 2: Vision-Text Alignment}  
Following the CLIP-style alignment procedure in DINO.txt, we freeze the trained ConvNeXt blocks (snowflake icons in Fig.~\ref{fig:train_nas_pipeline}, right), and train only the vision blocks that map visual features into the text embedding space. This stage uses a contrastive loss $\mathcal{L}_2 = \sum_i \mathcal{L}_{\text{contrast}}\big(\text{Text}(t), \text{Vision}_i(x)\big)$.
This two-stage pipeline ensures that the NAS supernet is compatible with CLIP-style open-vocabulary inference.
Neither stage is tied to this pair: conceptually, any distillable VFM teacher and CLIP-style alignment objective fit the same pipeline, with DINOv2 and DINO.txt the instance we build and evaluate.

\myparagraph{Stage 3: Selector Training}
The retention predictor of Alg.~\ref{alg:selection} is trained offline under leave-scene-out cross-validation.
We measure each subnet's accuracy once per scene, then regress the retention ratio $\mathrm{Acc}(M,C)/\mathrm{MaxAcc}$ from the class-set statistics $d$ and a normalized capacity descriptor of $M$ (FLOPs, width, depth, parameters), weighting each scene by its image count since retention measured on few images is noisy.
As tree ensembles pull predictions toward the mean, we calibrate the outputs with isotonic regression on a held-out split, so a fixed tolerance $\alpha$ still separates easy from hard tasks at inference.
This stage is one-time and negligible in cost next to Stages 1-2 (\suppref{sec:supp_selector_details}).

\myparagraph{Sandwich Sampling for Supernet Optimization}
To efficiently train the NAS search space, we follow the sandwich sampling rule~\cite{yu2020bignas}. In each iteration, we sample the \emph{largest} and \emph{smallest} subnets---corresponding to the two extremes in Tab.~\ref{tab:train_nas_space}---along with 2 additional subnets sampled uniformly from intermediate configurations. We compute gradients for all sampled subnets and average them to update shared weights, letting one supernet support a wide range of subnets with minimal accuracy gaps.

\myparagraph{Training Details}
For backbone distillation, we follow DINOv2 within our NAS framework, using the LVD-142M dataset~\cite{oquab2024dinov2}. We train the supernet for 625k iterations with a batch size of 4096 and a learning rate of 0.0002.  
For vision--text alignment, we adopt the setup of DINO.txt and train on the MetaCLIP dataset~\cite{xu2024demystifying} under the same NAS framework. This stage is trained for 100k iterations with a batch size of 4096 and a learning rate of 0.0004.
For the selector, we train 400 iterations with a learning rate of 0.05 and 15 leaves per tree.
Supernet training is a one-time cloud-side cost (\suppref{sec:supp_training}), amortized across deployments. All subnets share one checkpoint, so a device stores a single model.

\begin{table}
\vspace{-0.5em}
\centering
    {\small %
    \begin{tabular}{|c|ccc|}
    \hline
    Block & Dim. & Depth & Stride \\
    \hline
    Downsample-1     & $48\sim96$   & $1$       & $4$ \\
    ConvNext-Block-1 & $48\sim96$   & $3$       & $1$ \\
    \hdashline[2pt/5pt]
    Downsample-2     & $96\sim192$  & $1$       & $2$ \\
    ConvNext-Block-2 & $96\sim192$  & $3$       & $1$ \\
    \hdashline[2pt/5pt]
    Downsample-3     & $192\sim384$ & $1$       & $2$ \\
    ConvNext-Block-3 & $192\sim384$ & $9\sim27$ & $1$ \\
    \hdashline[2pt/5pt]
    Downsample-4     & $384\sim768$ & $1$       & $2$ \\
    ConvNext-Block-4 & $384\sim768$ & $3$       & $1$ \\
    \hline
    \#params & \multicolumn{3}{c|}{$6.2$M $\sim52.3$M} \\
    FLOPs & \multicolumn{3}{c|}{$1.29$G $\sim9.28$G} \\
    \hline
    \end{tabular}
    }
    \vspace{-0.3em}
\caption{Our NAS search space for the vision backbone. Depth denotes the number of layers and Dim. the dimensionality of each block.}
\vspace{-1em}
\label{tab:train_nas_space}
\end{table}

\section{Evaluation}
\label{sec:eval}

\subsection{Experimental Setup for Inference}
\label{subsec:eval_setup}

We evaluate zero-shot classification and open-vocabulary segmentation, two core tasks of contextual AIs.

\myparagraph{Zero-Shot Classification}
We evaluate on ImageNet-1K (IN1K)~\cite{deng2009imagenet}, Food101~\cite{bossard2014food101}, DTD~\cite{cimpoi14dtd} and Pets, following the inference procedure described in DINO.txt (\S\ref{subsec:method_adaptive_vfm}).

\myparagraph{Open-Vocabulary Segmentation}
We use ADE20K.
The effectiveness metric is mean Intersection-over-Union (mIoU).

\myparagraph{Metrics for Efficiency}
We use floating-point operations (FLOPs) and parameter count (\#params) to quantify the edge \VisionTower{'s} computational cost and memory usage (cloud-side per-update costs: \S\ref{subsec:method_manager}).

In addition, we acquire \textit{real-world} latency and energy usage as additional efficiency metrics, by deploying models on a test silicon featuring an ARM Ethos-U55 NPU~\cite{u55} fabricated in 7nm FinFET, a common industry setup (e.g.,~\cite{skillman2020technical,zhao2025h4h1}).
See \suppref{sec:supp_hardware} for hardware details.
The reported results under these metrics in \suppref{sec:supp_e2e} include averaged measured overhead of runtime wrapper, subnet switching, and other related costs.

\myparagraph{Evaluated NAS Subnets}
From the wide-spread NAS search space, we select five representative subnets described in Tab.\ref{tab:eval_subnets}.
\TinyNAS, \SmallNAS and \LargeNAS correspond to ConvNeXt-v2-N, -T and -S, respectively, while the remaining subnets are interpolated to span the entire search space.

\begin{table*}[t]
\centering
{\small %
    \begin{tabular}{|c|cc|cc|cc|}
    \hline
    Subnet & Depths & Dimensions & \#params & FLOPs & Latency & Energy \\
    \hline
    \MinNAS & [3,3,9,3] & [48,96,192,384] & 6.2M & 1.29G & $25$ms & $1.2$mJ \\
    \TinyNAS & [3,3,9,3] & [72,144,288,576] & 16.1M & 2.86G & $52$ms & $2.4$mJ \\
    \SmallNAS & [3,3,9,3] & [96,192,384,768] & 28.7M & 5.05G & $89$ms & $4.1$mJ \\
    \BaseNAS & [3,3,15,3] & [96,192,384,768] & 35.9M & 6.46G & $120$ms & $5.5$mJ \\
    \LargeNAS & [3,3,27,3] & [96,192,384,768] & 50.3M & 9.28G & $182$ms & $8.4$mJ \\
    \hline
    \end{tabular}
}
    \vspace{-0.5em}
\caption{Five representative NAS subnets selected from the search space
with sizes matching ConvNeXt-v2 variants (-N/-T/-S).}
\label{tab:eval_subnets}
\vspace{-1em}
\end{table*}

\subsection{Comparison with State-of-the-Art}
\label{subsec:eval_sota}

\myparagraph{Zero-Shot Classification}
We compare \ourname{} with state-of-the-art static and dynamic/adaptive baselines in Tab.\ref{tab:eval_zeroshot}.  
The static baselines are compared in the recent NoLA~\cite{imam2024clip}, and we use its settings.
The dynamic/adaptive baselines are reported by PatchRank~\cite{wu2025patchranking}.
Our approach improves over the comparably sized and larger static baselines and all reported dynamic baselines: \LargeNAS{} (50M) gains at least 7.9\% IN1K acc@1 over the 86M static models and 13.8\% over the 86M dynamic/adaptive ones.
The only exceptions are OpenCLIP~\cite{ilharco2021openclip} and DINO.txt, where the performance gap is primarily due to \textbf{\textit{model size differences}} (50M \LargeNAS{} vs.\ 304M OpenCLIP/DINO.txt).
Scene/context information is excluded in Tab.~\ref{tab:eval_zeroshot} as it is not suitable for object-oriented datasets.
As published baselines use their own pretraining data and objectives, Tab.~\ref{tab:eval_zeroshot} and Fig.~\ref{fig:intro_main_result} characterize end-to-end deployment trade-offs, while \S\ref{sec:abl} isolates each of our components under a fixed backbone and semantic pipeline.

Our models are also substantially more efficient in execution: the ViT-based baselines rely on an 86M ViT-B (423\,ms, 19.9\,mJ) or a 304M ViT-L (1119\,ms, 52.8\,mJ), at least $2.3\times$ the latency and $2.4\times$ the energy of our subnets (Tab.~\ref{tab:eval_subnets}).

\myparagraph{Open-Vocabulary Segmentation}
Beyond segmentation accuracy, we evaluate the accuracy-efficiency trade-off of our design against baselines.
Fig.~\ref{fig:intro_main_result} presents an end-to-end comparison, showing that \ourname achieves clear gains in the accuracy-efficiency (FLOPs) trade-off.
Accuracy-latency/energy trade-offs show similar trends, which are given in \suppref{sec:supp_e2e}.

\begin{table}
\centering
{\small\setlength{\tabcolsep}{2.5pt}%
\begin{tabular}{|@{}c@{}|c|c|cccc|}
\hline
& Model & \#param & Food101 & DTD & Pets & IN1K \\
\hline
\multirow{5}{*}{\rotatebox[origin=c]{90}{\textbf{\ourname}}}
 & \LargeNAS & 50M & 81.7 & 58.3 & 93.3 & 73.3 \\
 & \BaseNAS  & 36M & 80.5 & 57.9 & 93.3 & 72.4 \\
 & \SmallNAS & 29M & 79.2 & 57.2 & 92.8 & 71.3 \\
 & \TinyNAS  & 16M & 78.3 & 56.3 & 92.1 & 69.8 \\
 & \MinNAS   & 6M  & 73.8 & 54.1 & 91.4 & 65.5 \\
\hline
\multirow{9}{*}{\rotatebox[origin=c]{90}{Static}}
 & DINO.txt~\shortcite{jose2025dinov2}     & 304M & 93.4 & 67.5 & 95.3 & 81.6 \\
 & OpenCLIP~\shortcite{ilharco2021openclip}& 304M & -    & 55.6 & 83.9 & 74.0 \\
 & CLIP~\shortcite{radford2021CLIP}        & 86M  & -    & 42.9 & 85.0 & 61.9 \\
 & Waffle~\shortcite{roth2023waffling}     & 86M  & -    & 51.0 & 88.1 & 63.5 \\
 & LaFTer~\shortcite{mirza2023lafter}      & 86M  & -    & 46.1 & 82.7 & 64.2 \\
 & ProText~\shortcite{khattak2025learning} & 86M  & -    & 50.7 & 89.0 & 64.9 \\
 & NoLA~\shortcite{imam2024clip}           & 86M  & -    & 56.1 & 89.3 & 65.4 \\
 & TinyCLIP~\shortcite{wu2023tinyclip}
                                           & 45M & -    & -    & -    & 62.7 \\
 & MobileCLIP~\shortcite{vasu2024mobileclip}
                                           & 11M & -    & -    & -    & 67.8\\
\hline
\multirow{5}{*}{\rotatebox[origin=c]{90}{\quad Dynamic}}
 & EViT~\shortcite{liang2022evit}       & 86M & 80.3 & 43.3 & 87.1 & 58.1 \\
 & A-ViT~\shortcite{yin2022avit}      & 86M & 82.3 & 43.5 & 83.2 & 57.6 \\
 & ToMe~\shortcite{bolya2023tome}       & 86M & 82.4 & 41.5 & 87.2 & 58.3 \\
 & PatchRank~\shortcite{wu2025patchranking}  & 86M & 84.0 & 44.3 & 87.7 & 59.5 \\
\hline
\end{tabular}
}
\vspace{-0.5em}
\caption{Zero-shot classification. \ourname{} \textit{\textbf{outperforms}} the comparably sized (45--86M) and larger static baselines, and all reported 86M dynamic baselines, by at least 7.9\% IN1K acc@1. OpenCLIP and DINO.txt are much larger (304M vs.\ our $\leq$50M) and MobileCLIP (11M) much smaller. Baselines use their published pretraining setups.}
\label{tab:eval_zeroshot}
\vspace{-0.8em}
\end{table}

\section{Ablation and Analysis}
\label{sec:abl}

\ourname{'s} accuracy-efficiency gains come from three sources, as shown in Fig.\ref{fig:abl_manager}: a better-trained vision backbone (red curve vs the substantially weaker prior standalone models in Fig.\ref{fig:intro_main_result}), LLM-based runtime scene understanding that improves open-vocabulary accuracy (blue vs red), and runtime selection of efficient subnets (black vs blue).
Each comparison below changes one component at a time, so the benefit of each source is measured with the others held fixed.

Our two runtime mechanisms read only class names and subnet capacities, not backbone internals, so in principle they compose on any VFM backbone---a conceptual generalization we do not evaluate: a stronger one would enter as a better red curve, with filtering and selection trends unchanged.

\begin{figure}[t]
    \centering
    \includegraphics[width=0.66\linewidth]{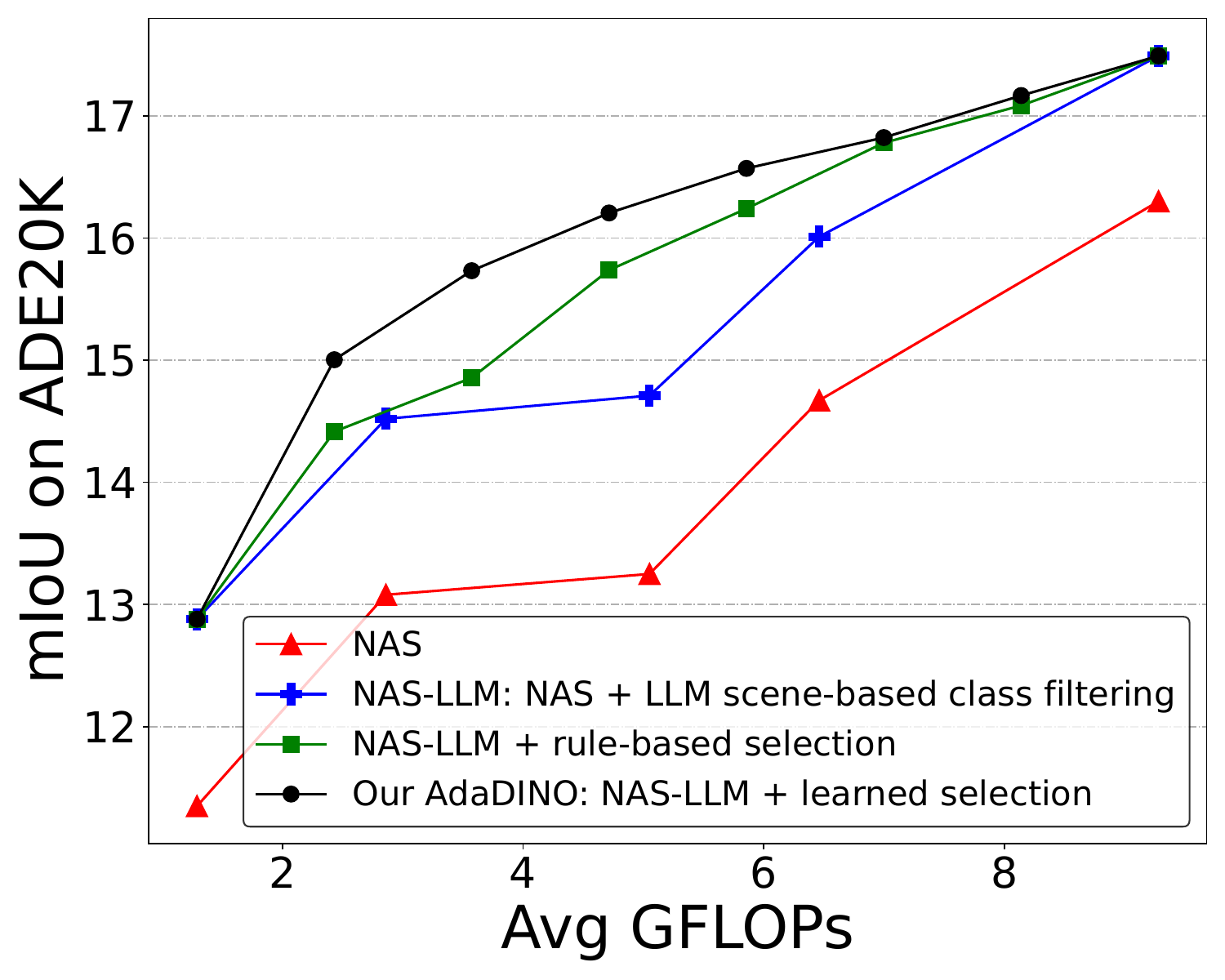}
    \vspace{-0.5em}
    \caption{Ablation of accuracy-efficiency gains. Red: our vision backbone with one globally fixed subnet per point and no semantic context; blue: the same family plus LLM-based class filtering, still with fixed subnets; green: the same semantic pipeline with a heuristic rule-based selector; black: with our learned selector. Green and black isolate the selector, sharing the same model family and semantic inputs (\S\ref{subsec:abl_selection}).}
    \label{fig:abl_manager}
    \vspace{-0.8em}
\end{figure}

\subsection{Stronger NAS-Enabled Vision Backbone}

Compared with DINO.txt-based standalone-distilled models, our NAS-based subnets achieve near-identical accuracy---e.g., \LargeNAS{} reaches 78.8\% on IN1K versus 79.1\% for its standalone counterpart, and stays within 0.9\% of the standalone model across all ten datasets---so weight sharing preserves the accuracy of independently trained models.
Full details are in \suppref{tab:supp_nas_backbone}.

\subsection{LLM-Assisted Runtime Scene Understanding}
\label{subsec:abl_semantic}
An LLM-based manager on the cloud assists \ourname{} by providing semantic understanding to improve the \TextTower.
We evaluate this process by first generating scene annotations on ADE20K with a LLM, then using a LLM to \textit{\textbf{filter out}} impossible class names and \textit{\textbf{resolve ambiguity}} by selecting contextually proper terms. The refined class set is provided to \ourname for open-vocabulary segmentation.
See \suppref{sec:supp_llm} for implementation of the cloud-side manager.

\myparagraph{Scene Annotation Generation}
We compare LLM-generated scene annotations with the ground-truth labels in ADE20K.
As in Tab.~\ref{tab:abl_scene_annotation}, a sufficiently powerful LLM such as Llama4~\cite{meta2024llama4} clearly outperforms the vanilla setting where no scene generation and class filtering is used.
Moreover, one concise annotation per scene approximates the accuracy of ground-truth scene labeling in ADE20K.

Another observation is that \textit{smaller} models \textit{benefit more} from the semantic filtering, with up to 1.53 mIoU improvement---a good fit for the always-on edge setting.

\myparagraph{Class Name Filtering}
Similarly, we ablate different commercial LLMs, prompts and implementation choices for class filtering in \suppref{subsec:supp_filter_ablation}.
LLM-based filtering yields substantial segmentation gains for \ourname---up to 1.76 mIoU over prior baselines and 1.53 mIoU over non-LLM Vanilla---and a visualization there shows more regular, coherent, semantically consistent masks.

\myparagraph{Out-of-Distribution Scenes}
Open-vocabulary tasks often encounter scenes outside any predefined training set, which calls for the \textit{open-ended} semantic reasoning of an LLM-based manager.
An OOD ablation in \suppref{subsec:supp_ood_scenes} shows that our open-set scene classifier remains robust while closed-set ones degrade significantly as fewer runtime scenes are known in advance.

\begin{table}[t]
\centering
{\small\setlength{\tabcolsep}{3pt}%
\begin{tabular}{|l|cc|cc|}
\hline
 & Llama4 & Llama3.2 & GT & Vanilla \\
\hline
\MinNAS   & 12.88 (+\underline{1.53}) & 11.7  & 13.11 & 11.35 \\
\TinyNAS  & 14.52 (+\underline{1.44}) & 13.68 & 14.54 & 13.08 \\
\SmallNAS & 14.71 (+\underline{1.46}) & 13.83 & 14.71 & 13.25 \\
\BaseNAS  & 16.01 (+\underline{1.34}) & 15.29 & 16.13 & 14.67 \\
\LargeNAS & 17.49 (+\underline{1.19}) & 16.9  & 17.53 & 16.30 \\
\hline
\end{tabular}%
}
\vspace{-0.5em}
\caption{
ADE20K mIoU under different \textit{scene annotation} strategies. First two columns use LLM-generated labels at runtime. GT uses ground-truth labels. Vanilla stands for not using scene for class filtering (red curve in Fig.\ref{fig:abl_manager}). Differences over Vanilla are shown in parentheses for Llama4, the scene annotator in our final design (blue in Fig.\ref{fig:abl_manager}).
LLM-based semantic understanding \defn{markedly improves} over Vanilla and \textbf{\textit{approximates}} ground-truth scenes.
}
\label{tab:abl_scene_annotation}
\vspace{-0.3em}
\end{table}

\subsection{Runtime Subnet Selection}
\label{subsec:abl_selection}
Fig.\ref{fig:abl_manager} studies how the cloud-side manager selects an edge-side VFM subnet by task difficulty, comparing our learned selector (black, \S\ref{subsec:method_manager}) with a heuristic rule-based selector (green, \suppref{sec:supp_rule_selector}) and a fixed-subnet baseline without runtime selection (blue).
All other parts are kept identical across the three settings.

Runtime subnet selection \textbf{\textit{substantially improves}} the accuracy--efficiency trade-off over using fixed subnets: at 15.0 mIoU, our learned selector reduces compute from 5.4 to 3.4 GFLOPs ($37\%$), while the rule-based selector requires 4.2 GFLOPs.
The learned selector also \textbf{\textit{consistently outperforms}} the heuristic in the intermediate operating range.
Its prediction error, per-tolerance violation rates, and cost against an oracle selector are quantified in \suppref{sec:supp_selector_details}.

\smallskip\noindent{\bf{Which Subnet is Selected?}}
Fig.\ref{fig:abl_subnet_selection} shows the fraction of selected subnets in ADE20K under different tolerance $\alpha$, optimized for FLOPs usage.
Even with strict accuracy requirements ($\alpha=0.98$), some scenes still select smaller subnets.
This necessitates \textbf{\textit{adaptive runtime selection}}, where certain task contexts can be efficiently handled by smaller models.

\begin{figure}[t]
    \centering
    \includegraphics[width=\linewidth]{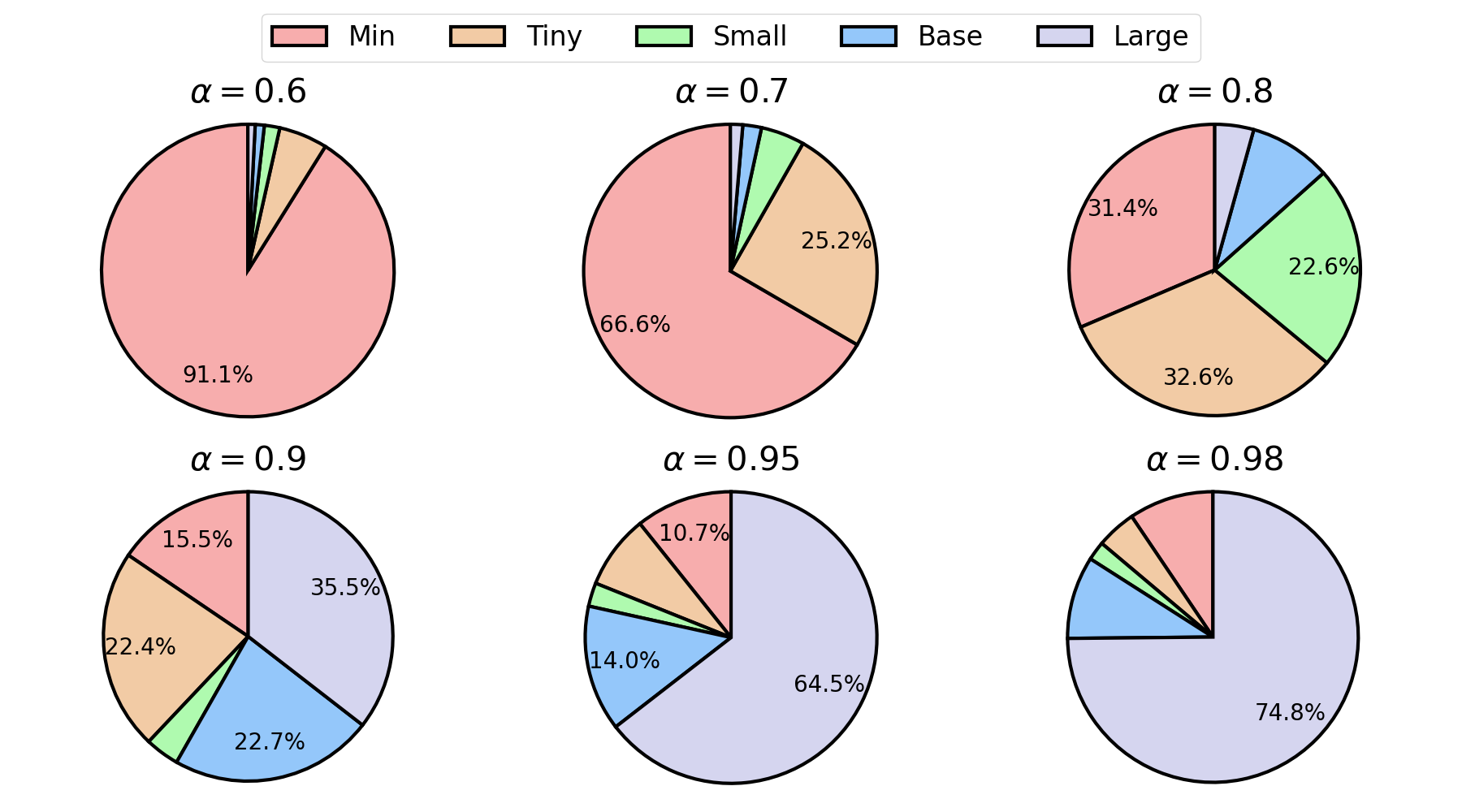}
    \vspace{-0.3em}
    \caption{Subnet selection across $\alpha$. Larger $\alpha$ imposes stricter accuracy requirements, leading to selection of larger subnets.}
    \label{fig:abl_subnet_selection}
    \vspace{-1em}
\end{figure}

\section{Conclusion}
\label{sec:concl}
We present \ourname{}, a context-adaptive framework for efficient edge DINOv2-distilled VFMs that combines a NAS-enabled model family with an LLM-assisted runtime manager to refine the active vocabulary and select the least-cost subnet by task context.
With backbone and semantic pipeline held fixed, learned selection alone cuts average compute by 37\%, and overall it improves the accuracy-efficiency trade-off by up to 5.2\% mIoU or 74.9\% fewer FLOPs---task-level adaptation built on standard, well-optimized components that stay deployable on edge accelerators.
Next steps include context-refresh scheduling and alternative context providers.

\bibliography{ref}
\clearpage
\renewcommand{\thesection}{\Alph{section}}
\setcounter{section}{0}

\section{Architecture of Basic Blocks}
\label{sec:supp_architecture}

\subsection{ConvNext-v2 Blocks}

We use ConvNeXt-v2~\cite{woo2023convnextv2} with \textbf{\textit{selective}} capacity as the core building block of our model, as shown in Fig.~\ref{fig:supp_convnext}.
All block widths ($dims$) are selectable during training and runtime, enabling the \textbf{\textit{adaptive}} behavior of our model.
We also replace GELU with ReLU for better compatibility and efficiency on edge devices.

\subsection{Downsample Layers}
We provide the basic structures of the downsampling layers we use in Fig.~\ref{fig:supp_downsample}.
As with the ConvNeXt blocks, the block widths ($dims$) are \textbf{\textit{selectable}} and adjusted according to the chosen configuration of the surrounding ConvNeXt blocks.

Downsample Layer 1 uses two consecutive $3\times3$-Conv2D layers with stride 2, yielding an effective downsampling factor of 4.
Downsample Layer 2 uses a single $3\times3$-Conv2D layer with stride 2.
Downsample Layers 3 and 4 instead apply $1\times1$-Conv2D layers with stride 2.

\begin{figure}[t]
\centering
\begin{subfigure}{0.23\linewidth}
    \centering
    \includegraphics[width=\linewidth]{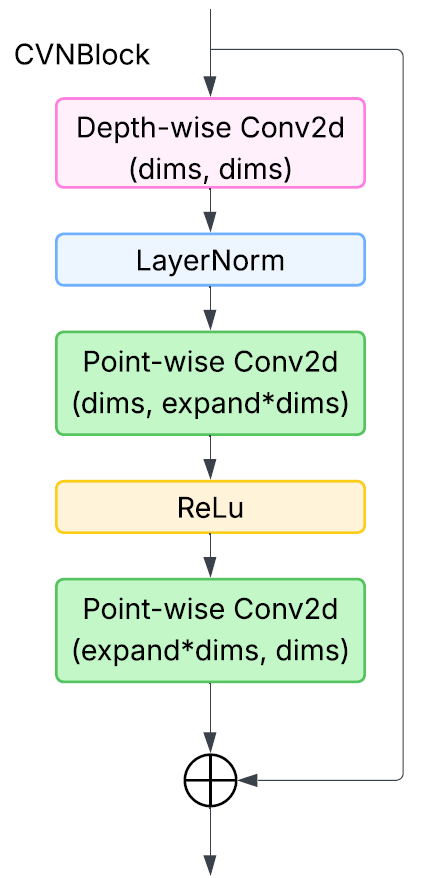}
    \phantomcaption
    \label{fig:supp_convnext}
\end{subfigure}
\hfill
\begin{subfigure}{0.75\linewidth}
    \centering
    \includegraphics[width=\linewidth]{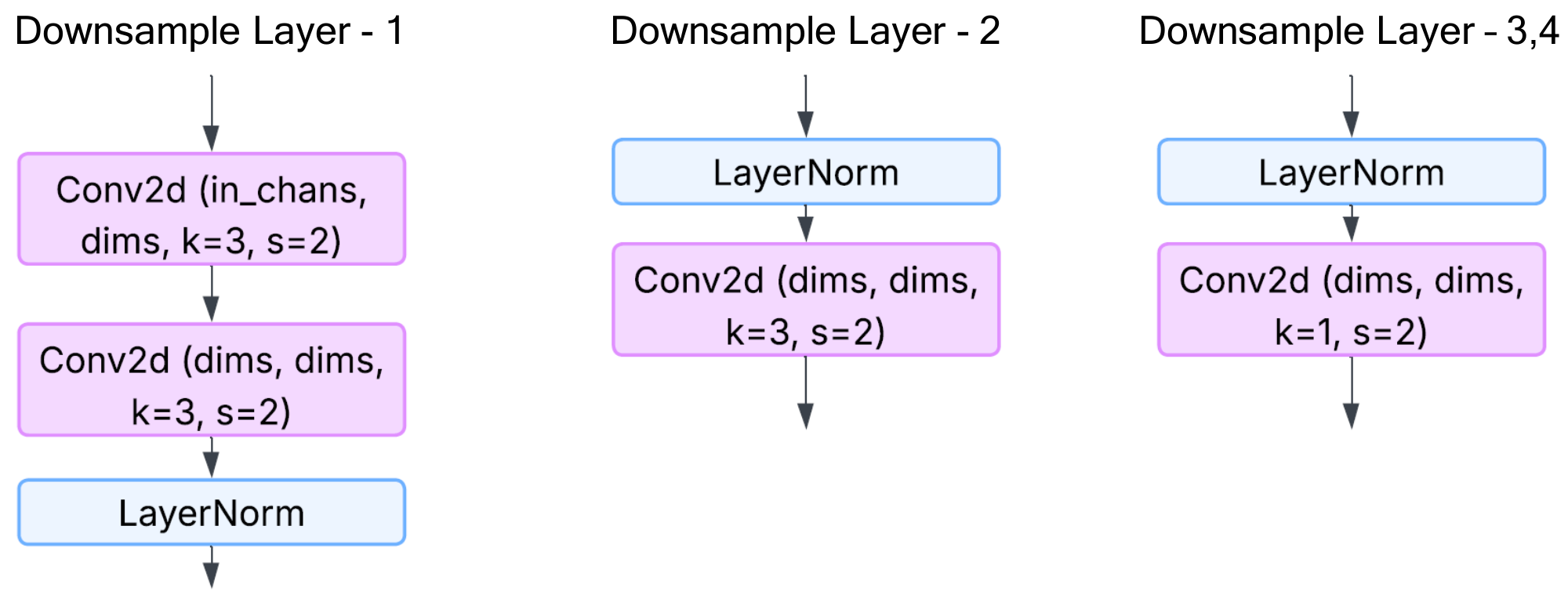}
    \phantomcaption
    \label{fig:supp_downsample}
\end{subfigure}
\vspace{-2em}
\caption{\textbf{(a) Left}: Selective ConvNeXt-v2 blocks. \textbf{(b) Right}: Downsample layers. The block widths ($dims$) are \textit{\textbf{selectable}} during training and runtime.}
\vspace{-1em}
\label{fig:supp_architecture}
\end{figure}

\section{Training Infrastructure}
\label{sec:supp_training}
Our models are trained on a cluster consisting of 48 AMD EPYC 7282 CPUs and 128 NVIDIA A100 PCIe GPUs, each with 80GB of GPU memory.
All reported experimental results, including both accuracy and efficiency measurements, are averaged over at least three independent runs.

\section{Hardware Platform and Evaluation Setup}
\label{sec:supp_hardware}
We adopt the ARM Ethos-U55~\cite{u55,skillman2020technical} as a representative edge Neural Processing Unit (NPU).
The test silicon (Fig.~\ref{fig:supp_evb}) is fabricated in 7\,nm FinFET and reflects the compute design of edge AR/VR devices such as Meta Ray-Ban~\cite{meta_rayban_smart_glasses}.
It operates at 560\,MHz, delivers $>$128\,GOP/s, occupies $\sim$7.8\,mm$^2$, and provides on-chip memory with 4\,GB/s aggregate bandwidth.

The device runs the Zephyr real-time operating system.
We deploy and evaluate various DNN layer types---standard, depth-wise, and point-wise convolutions; fully connected layers; attention and other transformer components; as well as activations and others---on the NPU using the ARM Ethos-U Vela toolchain.
The measured metrics of efficiency include execution latency and energy consumption of models.
The reported results in the main paper and supplementary material include averaged overheads of runtime wrappers, subnet switching, and other related costs.

\begin{figure}
    \centering
    \includegraphics[width=0.4\linewidth]{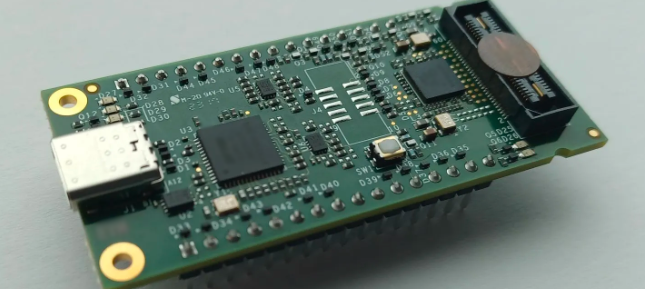}
    \caption{Our ARM Ethos-U55 test silicon.}
    \label{fig:supp_evb}
\end{figure}

\myparagraph{On-Device Memory Footprint}
Because every operating point is a subnet of one weight-sharing supernet, the device holds a single checkpoint rather than a separate model per size.
The supernet spans 6.2--52.3M parameters (\mainref{tab:train_nas_space}); at its largest it is about 52\,MB at 8-bit integer precision and 105\,MB at 16-bit---the two integer widths the Ethos-U55 datapath executes.
A smaller subnet activates only a subset of these weights, so parameter memory scales down with the selected width and depth, to roughly 6\,MB for \MinNAS{} at 8-bit.
Covering the same range of operating points with independent standalone backbones would instead cost the sum of their sizes, whereas the shared supernet keeps on-device storage at that of its single largest member while still exposing the full accuracy--efficiency frontier at runtime.
Activation and working-set memory depend on input resolution and the toolchain's buffer allocation and lie outside this parameter budget.

\myparagraph{Standard Components by Design}
\ourname{} is built from established, well-optimized components---a ConvNeXt-style convolutional backbone, weight-sharing NAS, foundation-model distillation, and a CLIP-style alignment recipe---rather than new operators. For an always-on edge target this is a deployment choice, not a gap in novelty. A fixed-function NPU such as the Ethos-U55 accelerates only the operators its compiler supports: the Vela toolchain maps supported layers to the NPU and leaves the rest to run on the far slower embedded CPU~\cite{u55,skillman2020technical}. This is a concrete case of the \emph{hardware lottery}, where an idea prevails largely by fitting the available hardware and tooling rather than on merit alone~\cite{hooker2021hardware}, so an operator without mature kernel and compiler support is handicapped whatever its accuracy. Theoretical cost only compounds this, since FLOPs is a loose proxy for on-device latency, which also turns on memory access and platform support~\cite{ma2018shufflenetv2}. Favoring standard, widely-optimized primitives keeps the whole model family deployable and fast on the target, whereas novel operators often fall back or fail to map and never ship. The contribution is accordingly not a new backbone but the task-context adaptation on top---semantic class filtering and learned subnet selection from the class vocabulary---which is independent of the backbone and improves as a stronger, equally deployable one becomes available.

\section{Per-Subnet Per-Task Accuracy of the NAS-Distilled Vision Backbone}
\label{sec:supp_backbone}
\Cref{tab:supp_nas_backbone} reports the accuracy of the five representative subnets drawn from our NAS-distilled vision backbone across ten vision datasets. Except IN1K (kNN), all datasets use logreg classifiers.
Compared with standalone-distilled models (last column), the largest NAS-based subnet (\LargeNAS) exhibits only \textbf{\textit{marginal}} drops---at most 0.9\% (on cars) and 0.3\% on IN1K---confirming that weight sharing across the supernet incurs negligible cost at the high-capacity end.
Sensitivity to model size, however, varies substantially across datasets: fine-grained tasks such as cars and Food101 degrade sharply for the smallest subnet (\MinNAS: $-18.4$\% and $-11.1$\%), whereas coarse-grained tasks such as Caltech-101 and Pets stay within a few points.
This heterogeneity is precisely what motivates adaptive, task-conditioned subnet selection at the edge rather than a single fixed model.

\begin{table*}[t]
\centering
    \begin{tabular}{|c|ccccc|c|}
    \hline
    & \multicolumn{5}{c|}{NAS-based Distillation} & Standalone \\
    Dataset & \MinNAS & \TinyNAS & \SmallNAS & \BaseNAS & \LargeNAS & \LargeNAS\\
    \hline
    IN1K~\cite{deng2009imagenet} & 69.4(-9.4) & 74.0 & 76.6 & 77.7 & 78.8(-0.3) & 79.1 \\
    Cal101~\cite{li2008cal101} & 95.2(-1.3) & 95.5 & 96.0 & 96.6 & 96.5(+0.3) & 96.2 \\
    Pets~\cite{parkhi2012pets} & 91.4(-3.3) & 92.9 & 93.9 & 94.6 & 94.7(+0.2) & 94.5 \\
    Cifar10~\cite{krizhevsky2009cifar} & 93.7(-4.4) & 95.8 & 97.1 & 97.9 & 98.1(-0.1) & 98.2 \\
    DTD~\cite{cimpoi14dtd} & 76.2(-4.8) & 79.4 & 80.4 & 80.7 & 81.0(-0.3) & 81.3 \\
    SUN397~\cite{xiao2010sun} & 69.7(-6.9) & 73.2 & 75.3 & 76.2 & 76.6(0.0) & 76.6 \\
    aircraft~\cite{maji2013finegrained} & 61.9(-9.0) & 66.1 & 68.9 & 69.8 & 70.9(+0.1) & 70.8 \\
    Cifar100~\cite{krizhevsky2009cifar} & 78.7(-10.2) & 83.1 & 86.1 & 87.9 & 88.9(-0.2) & 89.1 \\
    Food101~\cite{bossard2014food101} & 77.4(-11.1) & 82.6 & 86.1 & 87.5 & 88.5(-0.1) & 88.6 \\
    cars~\cite{gebru2017fine} & 66.8(-18.4) & 76.9 & 82.1 & 84.3 & 85.2(-0.9) & 86.1 \\
    \hline
    \end{tabular}
    \vspace{0.3em}
    \caption{Performance of our NAS-distilled vision backbone across vision tasks. Except IN1K (kNN), all other datasets use logreg classifiers. Compared with standalone-distilled models (last column), the NAS-based model shows only \textbf{\textit{marginal}} drops (differences in parenthesis of \LargeNAS). Sensitivity to model size varies across datasets (differences of \MinNAS), highlighting adaptive model selection based on task complexity.}
    \vspace{-0.5em}
\label{tab:supp_nas_backbone}
\end{table*}

\section{Granularity of User-Defined Classes is an Indicator of Task Difficulty}
\label{sec:supp_granularity}
In close-vocabulary tasks, the main sources of variation are determined by the datasets and workloads.
In contrast, for open-vocabulary tasks, an additional factor---how users define and execute the task at runtime---can influence task difficulty, alongside the underlying dataset distribution.

In this section, we study the impact of how users define \textbf{\textit{class names}} and their \textbf{\textit{granularities}}.
We conduct experiments on ImageNet-1K (IN1K)~\cite{deng2009imagenet} and ADE20K~\cite{zhou2019semantic}.
For IN1K, we evaluate two settings: the original 1000-class classification and a 7-class setting using very general class names.
The hierarchical grouping of class names is based on synsets in WordNet~\cite{fellbaum1998wordnet}.
For ADE20K, we also evaluate two settings: the original 150-class segmentation and a 9-class segmentation based on WordNet grouping with granularity similar to the 7-class IN1K.

To construct these coarse-grained settings, we group the IN1K categories into super-classes as follows.
For each class name, we first locate its entry in WordNet.
If it is absent, we use GPT-5 to map it to the closest WordNet term.
We then traverse the WordNet synset hierarchy and identify the lowest ancestor that contains $5\%\sim40\%$ of all IN1K classes, yielding the following set of seven super-classes.

\begin{tcolorbox}[title=Seven Super-classes of ImageNet-1K]
\label{box:supp_superclass_in1k}
1. animal\\
2. clothing\\
3. food\\
4. vehicle\\
5. instrumentality\\
6. structure\\
7. geological formation
\end{tcolorbox}

The ADE20K dataset~\cite{zhou2019semantic} is processed in a similar manner, but applies a heuristic class-name optimization from DINO.txt~\cite{jose2025dinov2} to improve segmentation performance, resulting in the following nine super-classes.

\begin{tcolorbox}[title=Nine Super-classes of ADE20K]
\label{box:supp_superclass_ade}
1. person;clothing\\
2. animal\\
3. plant\\
4. food\\
5. vehicle\\
6. nature;natural\_environment;geological\_formation\\
7. furniture;indoor\_furniture\\
8. road;path\\
9. building;structure
\end{tcolorbox}

As shown in \Cref{tab:supp_granularity},
while the original classification and segmentation tasks experience significant performance drops with smaller model sizes, the 7- or 9-class tasks with coarse-grained labels maintain strong performance even with smaller inference models.
These results demonstrate that user-defined class granularity has a significant impact on the difficulty of open-vocabulary tasks.
For simpler tasks, smaller models can achieve substantially higher efficiency while keeping nearly the same accuracy.
This observation is also visualized in \mainref{fig:intro_granularity}.

Consequently, NAS-enabled subnet selection is crucial for enabling efficient inference on edge devices across tasks of varying complexity.

\begin{table}
\centering
    \resizebox{\linewidth}{!}{
    \begin{tabular}{|c|ccccc|}
    \hline
    Dataset & \MinNAS & \TinyNAS & \SmallNAS & \BaseNAS & \LargeNAS \\
    \hline
    IN1K~\cite{deng2009imagenet} & 65.5 & 69.8 & 71.3 & 72.4 & 73.3 \\
    IN-WN7 & 73.2 & 73.8 & 74.1 & 74.1 & 74.3 \\
    \hline
    ADE~\cite{zhou2019semantic} & 11.35 & 13.08 & 13.25 & 14.67 & 16.30 \\
    ADE-WN9 & 19.01 & 19.62 & 19.38 & 20.23 & 20.22 \\
    \hline
    \end{tabular}
    }
    \vspace{1em}
\caption{Acc@1 on IN1K and mIoU on ADE20K. IN1K/ADE are original datasets, and IN-WN7/ADE-WN9 are 7-/9-class WordNet-based groupings. Coarser groupings show smaller accuracy drops as model size decreases, indicating lower task difficulty.}
\vspace{-0.5em}
\label{tab:supp_granularity}
\end{table}

The same effect also appears at the \textit{dataset} level.
\Cref{tab:intro_dino} reports zero-shot accuracy of the DINOv2 family across four classification datasets: simpler datasets such as Cal101 and Pets lose little accuracy with smaller models, whereas IN1K and Food101 degrade substantially, mirroring the granularity effect above.

\begin{table*}[t]
\centering
{\small
\begin{tabular}{|c|cc|c|c|c|c|}
\hline
DINOv2 & \#params & FLOPs & IN1K & Food101 & Cal101 & Pets \\
\hline
ViT-S & 22M & 4.6G & 79.0 (-4.4) & 89.1 (-5.6) & 97.0 (-0.6) & 95.1 (-1.6) \\
ViT-B & 86M & 17.6G & 82.1 (-1.3) & 92.8 (-1.9) & 96.1 (-1.5) & 96.2 (-0.5) \\
ViT-L & 304M & 61G & 83.5 (+0.1) & 94.3 (-0.4) & 97.5 (-0.1) & 96.6 (-0.1) \\
ViT-g & 1.8B & 700G & 83.4 & 94.7 & 97.6 & 96.7 \\
\hline
\end{tabular}
}
\caption{DINOv2 results. Numbers in parentheses show drops relative to ViT-g. Simple tasks (Cal101~\cite{li2008cal101}, Pets~\cite{parkhi2012pets}) exhibit small drops ($\leq1.6$) as model sizes decrease, while moderate/complex tasks (IN1K~\cite{deng2009imagenet}, Food101~\cite{bossard2014food101}) show larger drops (up to 5.6).}
\label{tab:intro_dino}
\end{table*}

\section{LLM-Based Runtime Scene Understanding}
\label{sec:supp_llm}

\subsection{Prompts for Runtime Scene Annotation Generation}
\label{subsec:supp_annotation_prompts}
For open-vocabulary segmentation on ADE20K~\cite{zhou2019semantic}, given an infrequently sampled image, we use the following prompts for Llama4 and Llama3.2~\cite{meta2024llama4} to generate a scene annotation.

\begin{tcolorbox}[title=Llama Prompt for Runtime Scene Annotation Generation]
\label{box:supp_prompt_llama}
Given an input image, imagine it comes from a computer vision benchmark such as ADE20K.\\
\\
1. Infer the single most appropriate scene or location category depicted in the image.\\
2. Use a WordNet-style noun phrase that is semantically aligned with ADE20K scene labels (e.g., "residential street", "indoor office", "natural park", "industrial warehouse").\\
3. Prefer generic, canonical scene names over narrative descriptions.\\
\\
Output format (mandatory):\\
Output only one phrase.\\
Do NOT use a code block, do NOT use markdown, and do NOT include quotes.\\
Only output the raw string itself (e.g., residential street).
\end{tcolorbox}

\subsection{Prompts for Runtime Scene-Aware Filtering}
\label{subsec:supp_filter_prompts}
For open-vocabulary segmentation on ADE20K~\cite{zhou2019semantic}, given a scene context, we use the following prompts for GPT-5~\cite{openai2025gpt5} and Gemini-2.5~\cite{google2025gemini25} to filter out implausible class names in the scene from the full set of 150 classes.

\begin{tcolorbox}[title=GPT-5 Prompt for Runtime Scene-Aware Filtering]
\label{box:supp_prompt_gpt}
Consider computer vision tasks in a scene of {\color{blue}\$scene}.
Given a scene description and a list of 150 object names, output a JSON dictionary 
where each key is an object name and the value is 0 if the object is highly unlikely in the scene, 
otherwise 1.
\\ \\
Requirements:\\
- Output only valid JSON (no explanations or text before/after).\\
- Keys must match exactly the provided object names.\\
- A total of 150 key-value pairs.\\
- Values must be integers 0 or 1 only.\\
\\
If multiple object names refer to ambiguous concepts, keep only one most likely to use when describing this particular scene.\\
For example:\\
- In a "sportsfield" scene, prefer "field" over "grass".\\
- In an outdoor scene, prefer "ground" over "floor".\\
- In an indoor scene, prefer "floor" over "ground".\\
- In an indoor scene, prefer "wall" over "building".\\
If there is no clear preference, you can keep both.\\
\\
Object names:
{\color{blue}\$object\_names}.
\\
Now output the JSON result only.
\end{tcolorbox}

\begin{tcolorbox}[title=Gemini-2.5 Prompt for Runtime Scene-Aware Filtering]
\label{box:supp_prompt_gemini}
You are a JSON-producing assistant.\\
Consider computer vision tasks in a scene of {\color{blue}\$scene}.
Given a scene description and a list of 150 object names, output a JSON dictionary 
where each key is an object name and the value is an integer:\\
- 1 if the object is likely or possible to be found in the scene.\\
- 0 if the object is highly unlikely or impossible in the scene.
\\ \\
Requirements:\\
- Starting with \{\{ and ending with \}\}. Do NOT use markdown or triple backticks.\\
- Output only valid JSON (no explanations or text before/after).\\
- Keys must match exactly the provided object names.\\
- A total of 150 key-value pairs.\\
- Values must be integers 0 or 1 only.\\
\\
If multiple object names refer to ambiguous concepts, keep only one most likely to use when describing this particular scene.\\
For example:\\
- In a "sportsfield" scene, prefer "field" over "grass".\\
- In an outdoor scene, prefer "ground" over "floor".\\
- In an indoor scene, prefer "floor" over "ground".\\
- In an indoor scene, prefer "wall" over "building".\\
If there is no clear preference, you can keep both.\\
\\
Object names:
{\color{blue}\$object\_names}.

Now output the JSON result only.
\end{tcolorbox}

The two prompts above generate an unbounded set of candidate classes based on the scene context.
We also study a method that restricts predictions to a fixed top-$k$ class subset, using the GPT-5 prompts listed below.

\begin{tcolorbox}[title=Top-$k$ Prompt for GPT-5 for Runtime Scene-Aware Filtering]
\label{box:supp_prompt_topk}
Consider computer vision tasks in a scene of {\color{blue}\$scene}.
Given a scene description and a list of 150 object names, output only the top \$\textbf{\textit{k}} most relevant objects likely to appear in this scene.
\\ \\
Requirements:\\
- Output only valid JSON (no explanations or text before/after).\\
- Keys must match exactly the provided object names.\\
- A total of 150 key-value pairs.\\
- Values must be integers 0 or 1 only.\\
\\
If multiple object names refer to ambiguous concepts, keep only one most likely to use when describing this particular scene.\\
For example:\\
- In a "sportsfield" scene, prefer "field" over "grass".\\
- In an outdoor scene, prefer "ground" over "floor".\\
- In an indoor scene, prefer "floor" over "ground".\\
- In an indoor scene, prefer "wall" over "building".\\
If there is no clear preference, you can keep both.\\
\\
Object names:
{\color{blue}\$object\_names}.
\\
Now output the JSON result only.
\end{tcolorbox}

\subsection{End-to-End Segmentation Results of Class Name Filtering}
\label{subsec:supp_filter_ablation}
As shown in \Cref{tab:abl_scene_filter}, incorporating LLM-based grouping and filtering of vocabulary yields substantial segmentation gains for \textit{both} our adaptive \ourname and non-adaptive baselines (e.g.,~\cite{jose2025dinov2}).
Our approach gains up to 1.76 mIoU over prior baselines and 1.53 mIoU over non-LLM Vanilla.
Fig.~\ref{fig:abl_llm_masks} further shows that LLM-based understanding yields more regular, coherent, and semantically consistent masks.

Among the commercial LLMs we evaluated, GPT-5~\cite{openai2025gpt5} achieves the best performance.
Using alternative LLMs (e.g., Gemini-2.5~\cite{google2025gemini25}) or smaller variants (GPT-5-mini) results in noticeable performance degradation.
The first three methods in Tab.~\ref{tab:abl_scene_filter} yield an indefinite number of candidate classes based on scene context.
The fourth method restricts predictions of GPT-5 to a fixed top-25 class subset, which leads to poorer performance.

\begin{table}[t]
\centering
\resizebox{\linewidth}{!}{%
\begin{tabular}{|l|ccc|c|}
\hline
 & GPT5 & GPT5-mini & Gemini2.5 & Top25 \\
\hline
\MinNAS   & 12.88 (+\underline{1.53}) & 12.75 & 12.57 & 13.35 \\
\TinyNAS  & 14.52 (+\underline{1.44}) & 14.28 & 14.19 & 14.38 \\
\SmallNAS & 14.71 (+\underline{1.46}) & 14.41 & 14.37 & 14.48 \\
\BaseNAS  & 16.01 (+\underline{1.34}) & 15.77 & 15.79 & 15.80 \\
\LargeNAS & 17.49 (+\underline{1.19}) & 17.13 & 17.20 & 16.82 \\
\hline
ViT-L     & 21.00 (+\underline{0.86}) & 20.68 & 20.68 & 19.91 \\
\hline
\end{tabular}%
}
\vspace{1em}
\caption{ADE20K~\cite{zhou2019semantic} mIoU with different LLM-based \textit{class filtering} methods.
Incorporating LLM \defn{notably improves} open-vocabulary segmentation for both \ourname{} and DINO.txt baseline~\cite{jose2025dinov2}, with \textit{smaller} models benefiting more. Differences over the non-LLM Vanilla setting are shown in parentheses for GPT-5.}
\label{tab:abl_scene_filter}
\end{table}

\begin{figure*}
    \centering
    \includegraphics[width=0.6\textwidth]{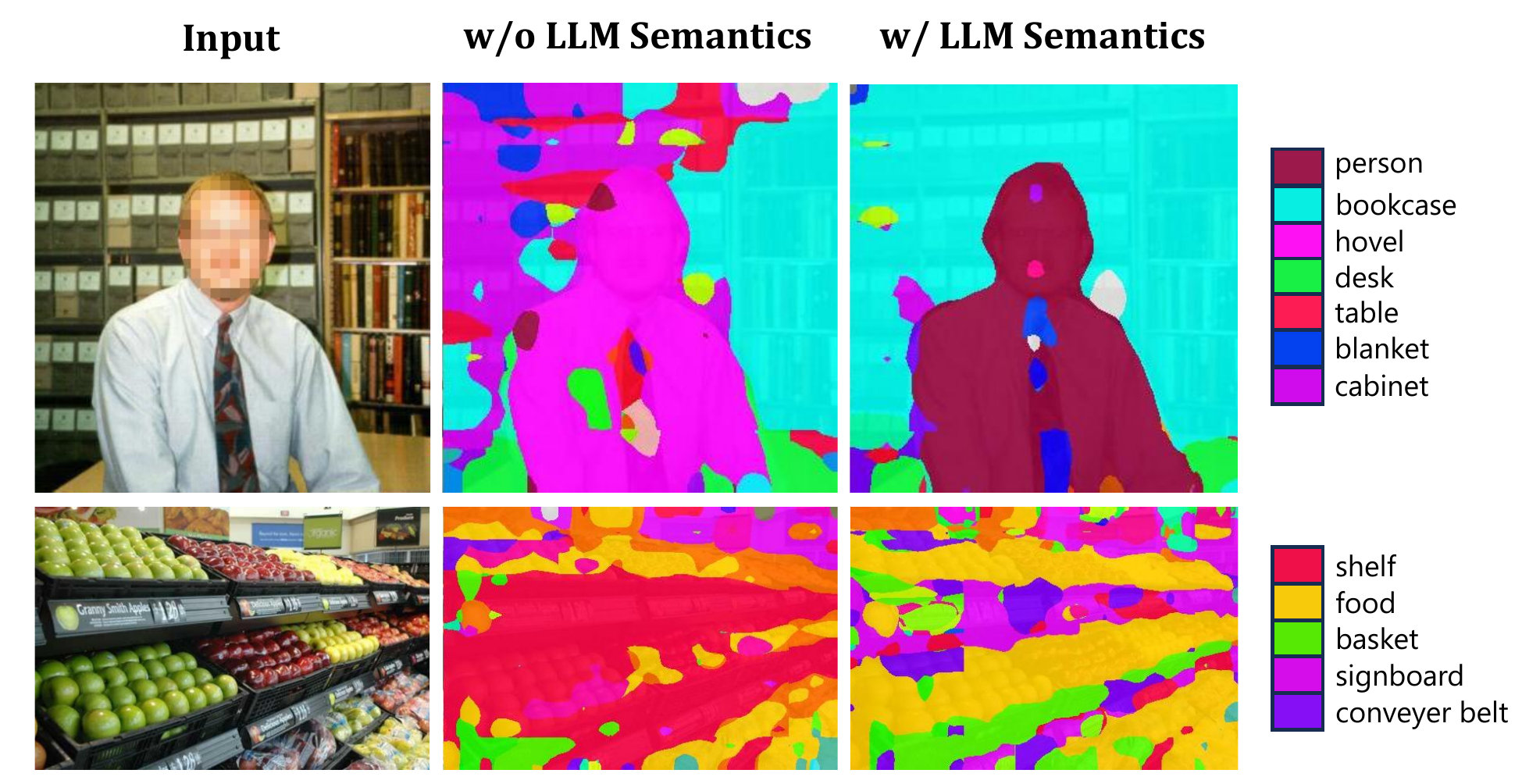}
    \vspace{-0.5em}
    \caption{Open-vocabulary segmentation on ADE20K~\cite{zhou2019semantic} with and without LLM-based scene understanding. LLM guidance produces more \textit{regular and coherent} masks.}
    \vspace{-0.7em}
    \label{fig:abl_llm_masks}
\end{figure*}

\subsection{Quality of Scene-Aware Class Filtering}
\label{subsec:supp_filter_quality}
\Cref{tab:supp_filter_quality} shows the average recall and precision for each LLM prompt used in scene-aware class name filtering in ADE20K~\cite{zhou2019semantic}.
The manager is designed as a conservative vocabulary filter rather than an object detector: removing a class that is actually present hurts far more than retaining an extra plausible one, so the prompts deliberately prioritize recall over precision.
Its benefit comes from removing semantically incompatible labels and resolving dataset-specific ambiguity, not from pinpointing exactly the objects in the image.
Top-$k$ methods exhibit high precision but low recall by being over-aggressive in filtering, while Gemini-2.5 achieves high recall but low precision by being over-conservative, leaving too many irrelevant classes remaining.
Both lead to poorer end-to-end performance in open-vocabulary segmentation compared to GPT-5, but still provide slight end-to-end improvements over the non-LLM baselines, such as DINO.txt~\cite{jose2025dinov2}.

\begin{table}[t]
\centering
\begin{tabular}{|c|cc|} %
\hline
Method & Recall & Precision \\
\hline
GPT-5 & 89.7 & 6.1 \\
Gemini-2.5 & 94.7 & 3.2 \\
\hline
Top-20 & 71.4 & 9.1 \\
Top-25 & 77.1 & 8.0 \\
\hline
\end{tabular}
\vspace{1em}
\caption{Average recall and precision for each LLM prompt used in scene-aware class filtering on ADE20K~\cite{zhou2019semantic}. Among these methods, GPT-5 achieves the best end-to-end performance for open-vocabulary segmentation.}
\label{tab:supp_filter_quality}
\end{table}

\subsection{LLM vs. Non-LLM Scene Detection under Out-of-Distribution Scenes}
\label{subsec:supp_ood_scenes}
The cloud-side manager of \ourname{} relies on an LLM throughout its semantic pipeline: it understands the current scene, generates an open-ended scene label (\Cref{subsec:supp_annotation_prompts}), and then filters class names conditioned on that scene (\Cref{subsec:supp_filter_prompts}).
A natural question is whether such an LLM is necessary, or whether a lightweight non-LLM scene classifier could fulfill the same role.
Non-LLM scene classifiers are efficient for closed-vocabulary settings, but \ourname{} targets \textbf{\textit{open-vocabulary}} tasks, where runtime scenes may be out-of-distribution (OOD) with respect to any predefined scene set.
Forcing them into a fixed scene set can mislead the downstream class filtering.

To validate this, we conduct an OOD open-vocabulary segmentation experiment on ADE20K~\cite{zhou2019semantic}.
The LLM-based method performs open-ended scene detection and class filtering, as in our final design.
For the non-LLM baselines, we use a DINOv2-based~\cite{oquab2024dinov2} scene classifier, but restrict its output to a subset of ADE20K scene labels, emulating deployments where only a fraction of runtime scenes are known in advance.
We construct this scene subset in two ways: \romenum{1} \textit{Random}, which randomly keeps a fraction of ADE20K scenes; and \romenum{2} \textit{WordNet-Sim}, which selects representative scene labels by clustering scene names with WordNet~\cite{fellbaum1998wordnet} similarity.
The classifier maps each image to one of these subset scene labels.
All non-LLM baselines are further given an \textbf{\textit{oracle}} lookup from the predicted retained scene to its filtered class set, making them strong heuristics.

As shown in \Cref{tab:supp_non_llm}, despite the oracle lookup, the non-LLM baselines still suffer noticeable drops compared with the LLM-based method, and the gap widens as fewer runtime scenes are known in advance.
These results support the need for the \textbf{\textit{flexible semantic grounding}} of LLMs in open-vocabulary tasks: an LLM-based scene understander generalizes to OOD scenes, whereas a closed-vocabulary supervised classifier misassigns them to its fixed scene set and propagates the error into class filtering.

\begin{table}[t]
\centering
\setlength{\tabcolsep}{2.5pt}
\begin{tabular}{|c|c|cc|}
\hline
Known scenes & \ourname{} (LLM) & Random & WordNet-Sim \\
\hline
80\% & 17.53 & 16.09 & 17.35 \\
70\% & 17.53 & 15.17 & 17.27 \\
60\% & 17.53 & 13.48 & 15.15 \\
40\% & 17.53 & 12.56 & 15.13 \\
\hline
\end{tabular}
\vspace{1em}
\caption{ADE20K~\cite{zhou2019semantic} mIoU of the \LargeNAS{} subnet with LLM-based open-ended scene detection versus non-LLM scene classifiers restricted to a known subset of scenes. Despite an oracle scene-to-class lookup, the non-LLM baselines suffer noticeable drops that grow as fewer runtime scenes are known in advance.}
\label{tab:supp_non_llm}
\end{table}

\section{Learned Selector: Implementation Details and Calibration}
\label{sec:supp_selector_details}
This section collects the implementation details of the learned selector (\mainref{alg:selection}) in one place and evaluates its retention predictor directly.

\myparagraph{Inputs}
The class names $C$ of the current task are embedded by the \TextTower{} into 2048-dimensional vectors, precomputed and fetched by lookup, so the selector runs no text-encoder pass of its own.
\textsc{SetStats} turns them into a 27-d descriptor $d$ that is invariant to the order and size of the class set.
For each of two prompt templates (``a photo of \textit{class}'' and the bare class name), the embeddings are L2-normalized and eleven statistics of their pairwise cosine-similarity matrix are taken: mean, standard deviation, min, max, and five percentiles (10/25/50/75/90) of the off-diagonal entries, plus the mean and max of the per-class nearest-neighbor similarity.
The class count and its log complete the descriptor, together with three schema fields that are constant across our segmentation tasks.
Class sets with fewer than two names have no pairs, so the pairwise statistics are zeroed and the size features carry the signal.
Computing $d$ averages 1\,ms per class set on a single CPU core.
Each candidate subnet $M$ is described by four capacity features---FLOPs, width, depth, and parameter count---each normalized by the largest subnet's value.

\myparagraph{Training Data and Labels}
Supervision comes from the ADE20K validation scenes: each validation image carries a scene-category label, and each scene's class set is its row in a human-verified scene-to-class table over the 150-class vocabulary.
Dropping scenes without a usable class set or largest-subnet measurement leaves 664 scenes covering 1992 of the 2000 validation images, or $664\times5=3320$ (scene, subnet) rows.
The label of a row is the retention ratio $\mathrm{Acc}(M,C)/\mathrm{MaxAcc}$, where accuracy is the scene's mIoU restricted to its class set and pooled over its images, and $\mathrm{MaxAcc}$ is the same measurement for the largest subnet.
Most scenes hold a single validation image (median 1, mean 3.0), so scene-level labels are noisy, and scenes are weighted by their image counts during fitting.

\myparagraph{Public Dataset Choice and Generalization}
ADE20K is, to our knowledge, the public segmentation benchmark that supplies all three ingredients this supervision needs at once: a scene-category taxonomy, dense masks over one shared vocabulary, and scene labels from which a human-verified scene-to-class table can be built.
Object- or stuff-centric benchmarks such as COCO-Stuff or Pascal Context carry no comparable scene taxonomy, so they cannot furnish per-scene retention labels without substantial extra annotation, and we run every reported selector study on ADE20K.
The mechanism, though, is not tied to this vocabulary: the descriptor reads only statistics of class-name text embeddings and per-subnet capacity, both defined for any class set, so a selector fit on one vocabulary transfers to another with no change to the model.
The settings this is ultimately meant for are in-house, open-vocabulary deployments whose data cannot be released. ADE20K is the public proxy on which the complete scene-to-class supervision happens to be available, and the held-out results below stand in for that target.

\myparagraph{Regression Model and Calibration}
A gradient-boosted regressor maps $(d, M)$ to the retention ratio and is constrained to be monotone in all four capacity features (400 iterations, learning rate 0.05, 15 leaves per tree).
Tree ensembles pull predictions toward the mean, which would blur the separation between easy and hard tasks, so the raw outputs are calibrated with isotonic regression fit on a held-out half of each training fold.
Every selector prediction reported in this paper is made under leave-scene-out cross-validation: scenes are grouped into five folds, and each scene is served by a model that never saw it, so the evaluated scenes and their class sets are unseen at fit time.

\myparagraph{Training vs.\ Deployment Inputs}
The supervision above is built entirely from ground truth: scenes come from the dataset's scene-category labels, and each class set is a human-verified scene-to-class row.
At inference, the selector instead receives what the runtime pipeline produces: a Llama4-generated scene annotation followed by GPT-5 class filtering (\mainref{subsec:abl_semantic}).
The deployed selector thus runs on LLM-generated, \emph{open-vocabulary} class sets that never appear in its fitting data, and the descriptor accepts them unchanged because it reads only statistics of class-name text embeddings.
The diagnostics below probe the predictor on held-out scenes under the ground-truth protocol, isolating predictor quality from LLM variability, while the end-to-end results in the main paper exercise the full LLM-driven path.

\myparagraph{Runtime Behavior}
The selector does not name a subnet directly.
It predicts a retention ratio for every subnet, and an explicit decision rule returns the least-cost subnet whose prediction meets $\alpha$, falling back to the largest subnet when none qualifies (\mainref{alg:selection}).
This separation lets one trained selector serve different operating tolerances without retraining.
The selector runs on the cloud once per context update, and its cost is negligible next to the MM-LLM call in the same update.

\myparagraph{Predictor Accuracy}
On scenes with reliable labels (at least three validation images), the deployed predictor reaches a mean absolute error of 0.086 on the retention ratio (RMSE 0.123).
Single-image scenes, whose measured retention is itself noise-dominated, inflate the raw error over all 3320 held-out rows to 0.198, still an improvement from 0.225 before calibration.
Binned by predicted retention, the well-populated bins below 0.9 are conservative---measured retention exceeds the prediction---while calibrated predictions saturating near 1.0 overshoot slightly.

\myparagraph{Meeting the Tolerance}
\Cref{tab:supp_selector_calib} reports selector behavior at the six tolerances of \mainref{fig:abl_subnet_selection}, with compute relative to always running the largest subnet.
Both the mean per-scene retention of the selected subnets and the dataset-level pooled retention track the requested $\alpha$ closely, the latter meeting the tolerance at 32 of the 34 operating points of the sweep behind \mainref{fig:abl_manager} (worst shortfall 0.008 at $\alpha=0.99$), while average compute stays 8.0--85.7\% below the largest subnet.
Because most scenes carry a single validation image, per-scene retention is measured under heavy label noise, so \mainref{alg:selection} is best read as selecting the cheapest subnet whose \emph{predicted} retention reaches $\alpha$: the tolerance is honored in expectation and in the dataset-level aggregate rather than as a per-scene certificate.
Turning it into a per-context high-probability guarantee---predicting a lower retention quantile and selecting against that instead of the mean---is a natural extension that our formulation admits and we leave to future work.

\myparagraph{Oracle Comparison}
We also compare against an oracle that reads each scene's measured retention and picks the cheapest qualifying subnet, incurring no violations by construction.
At the strict tolerances most relevant to the always-on setting ($\alpha\ge0.94$), the learned selector stays within 7.5\% of this oracle's average compute. The margin widens at looser tolerances, where the oracle most exploits single-image label noise.
Because the oracle reads the same few validation images that define the labels, it marks a lower bound on achievable compute rather than a reachable target.

\begin{table}[t]
\centering
{\small\setlength{\tabcolsep}{4pt}
\begin{tabular}{|c|c|c|c|}
\hline
$\alpha$ & scene ret. & pooled ret. & FLOPs (\%) \\
\hline
0.6  & 0.842 & 0.749 & $-$85.6 \\
0.7  & 0.840 & 0.763 & $-$84.0 \\
0.8  & 0.888 & 0.818 & $-$70.8 \\
0.9  & 0.931 & 0.939 & $-$29.3 \\
0.95 & 0.947 & 0.968 & $-$14.4 \\
0.98 & 0.970 & 0.982 & $-$8.8 \\
\hline
\end{tabular}
}
\vspace{-0.3em}
\caption{Selector behavior on the 664 held-out ADE20K scenes at the six tolerances of \mainref{fig:abl_subnet_selection}. Scene ret.: mean per-scene retention of the selected subnets; pooled ret.: dataset-level mIoU ratio over all images; FLOPs: average compute relative to always running the largest subnet (negative = savings).}
\label{tab:supp_selector_calib}
\end{table}

\section{Baseline: Rule-Based Model Selection}
\label{sec:supp_rule_selector}
We include a simple rule-based baseline only for comparison, which is not used in the final design of our \ourname.
The baseline assigns each task to a subnet using a fixed lookup rule based solely on the number of class names.
Tasks are ranked by class count and mapped to progressively larger subnets, while a single global parameter controls the overall allocation and traces an accuracy-efficiency frontier.
Since it uses neither text embeddings nor a learned predictor, the method is inexpensive and easy to implement.
However, this coarse heuristic ignores the identities and relationships of the classes, which is reflected as poorer performance in \mainref{fig:abl_manager}, motivating the more fine-grained learned selector used in \ourname{}.

\section{Accuracy-Efficiency Trade-offs with Additional Metrics}
\label{sec:supp_e2e}

\myparagraph{End-to-End Comparison}
In the main paper, we present the accuracy-efficiency trade-off (\mainref{fig:intro_main_result}) using FLOPs.
Here, we additionally report results based on average execution latency and energy consumption, respectively in Fig.~\ref{fig:supp_e2e_latency} and Fig.~\ref{fig:supp_e2e_energy}.
The compared baselines are identical to those in \mainref{fig:intro_main_result}~\cite{ghiasi2022scaling,jose2025dinov2,radford2021CLIP,fang2024data,ilharco2021openclip,xing2023rewrite,xu2024demystifying,xu2023learning,zhai2023sigmoid}.
They demonstrate that \ourname achieves significant gains in the accuracy-efficiency trade-off.
Since these metrics exhibit the same trends and lead to the same main conclusions, we include them only in the Supplementary Material.

\begin{figure}[t]
    \centering
    \includegraphics[width=\linewidth]{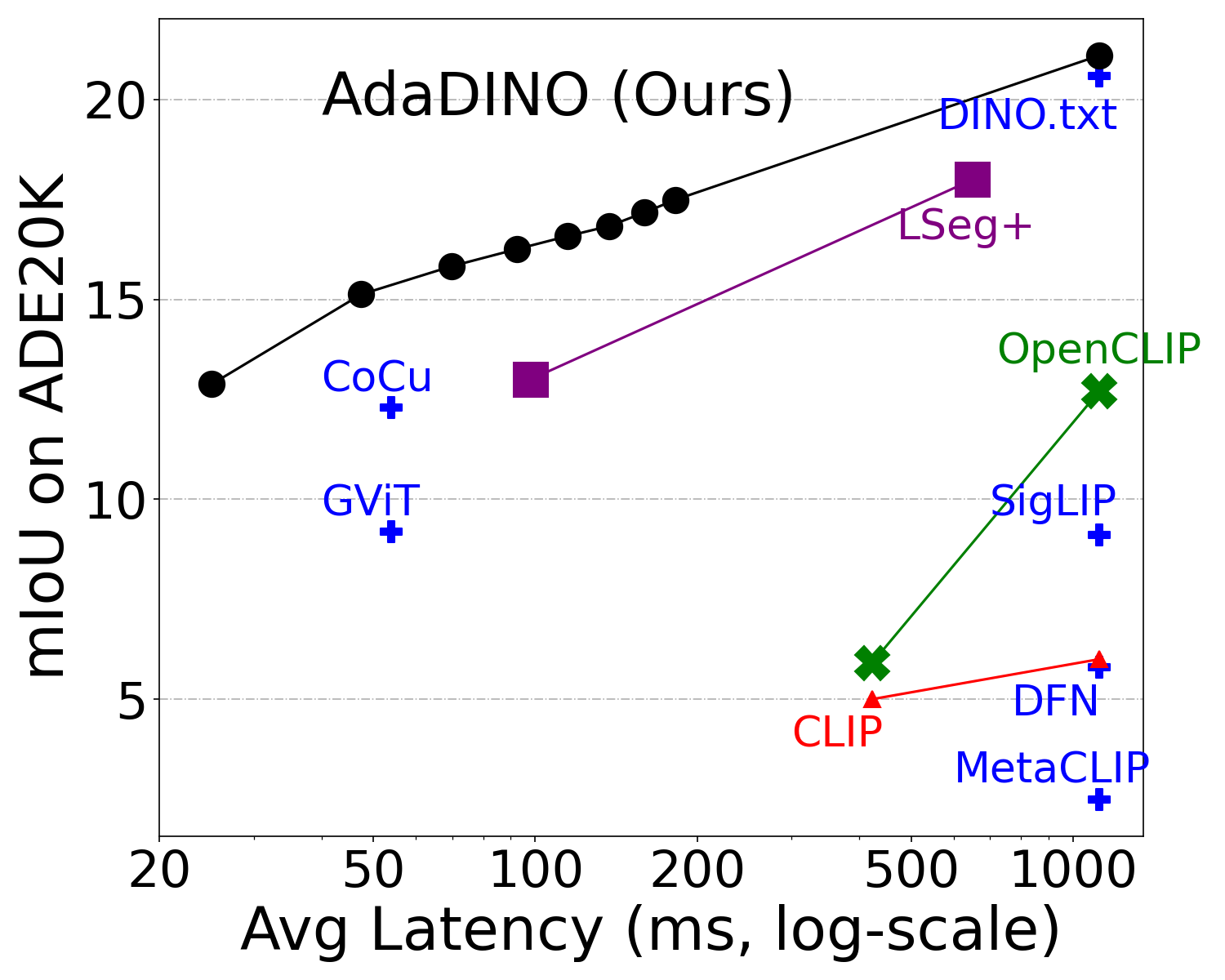}
    \vspace{-1.5em}
    \caption{End-to-end trade-off between mIoU on open-vocabulary ADE20K segmentation~\cite{zhou2019semantic} and average execution latency.}
    \label{fig:supp_e2e_latency}
\end{figure}

\begin{figure}[t]
    \centering
    \includegraphics[width=\linewidth]{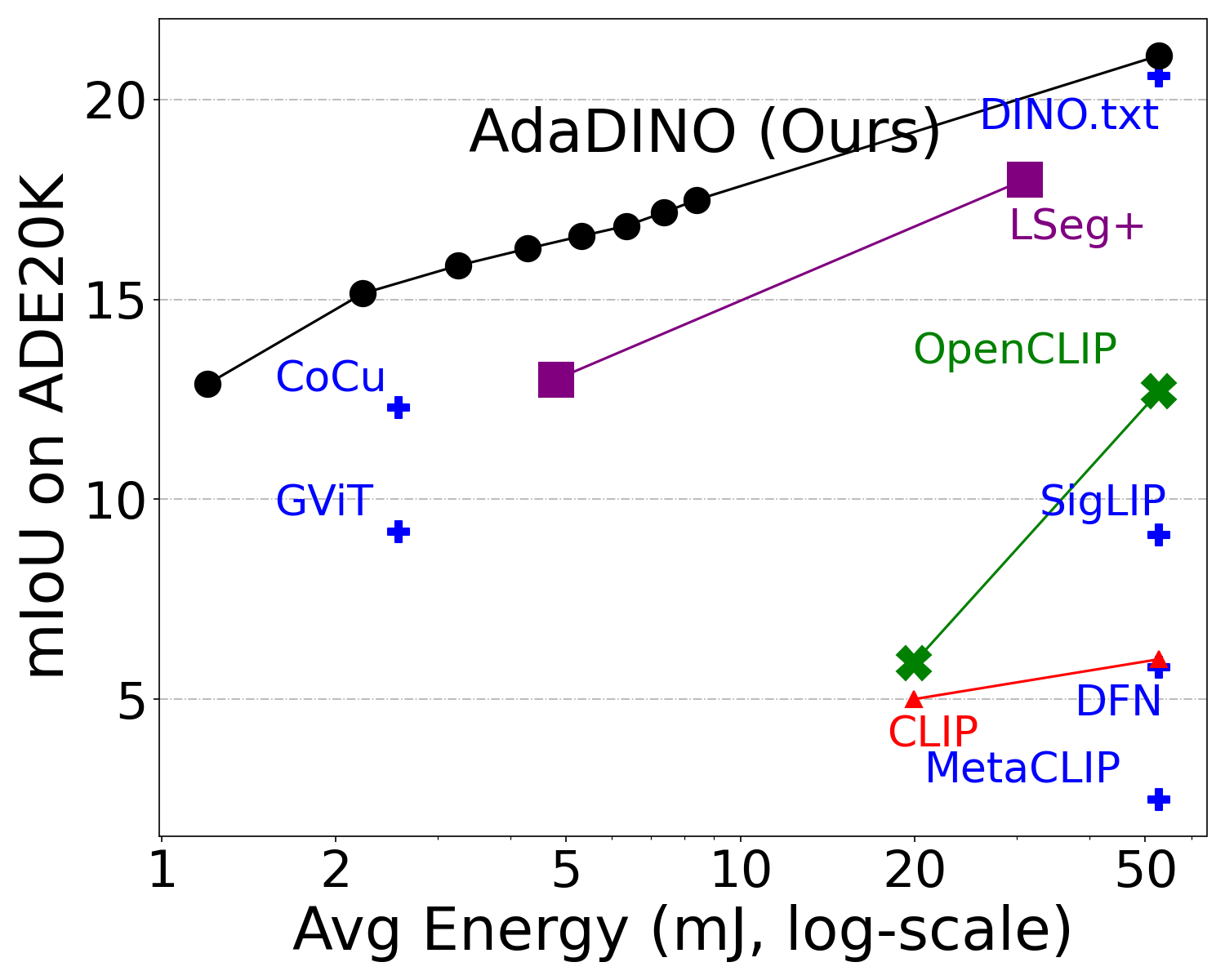}
    \vspace{-1.5em}
    \caption{End-to-end trade-off between mIoU on open-vocabulary ADE20K segmentation~\cite{zhou2019semantic} and average energy consumption.}
    \label{fig:supp_e2e_energy}
\end{figure}

\myparagraph{Ablation Across Additional Metrics}
In the main paper, we present the ablation of \ourname{'s} accuracy-efficiency gains (\mainref{fig:abl_manager}) using FLOPs.
Here, we additionally report the ablation based on average execution latency, energy consumption, and number of model parameters, respectively in Fig.~\ref{fig:supp_abl_latency}, Fig.~\ref{fig:supp_abl_energy}, and Fig.~\ref{fig:supp_abl_params}.
The compared settings are identical to those in \mainref{fig:abl_manager}, including the rule-based selection baseline described in \Cref{sec:supp_rule_selector}.
They demonstrate that each component of \ourname provides a clear benefit, with the learned selector consistently yielding the best trade-off.
Since these metrics exhibit the same trends and lead to the same main conclusions, we include them only in the Supplementary Material.

\begin{figure}[t]
    \centering
    \includegraphics[width=\linewidth]{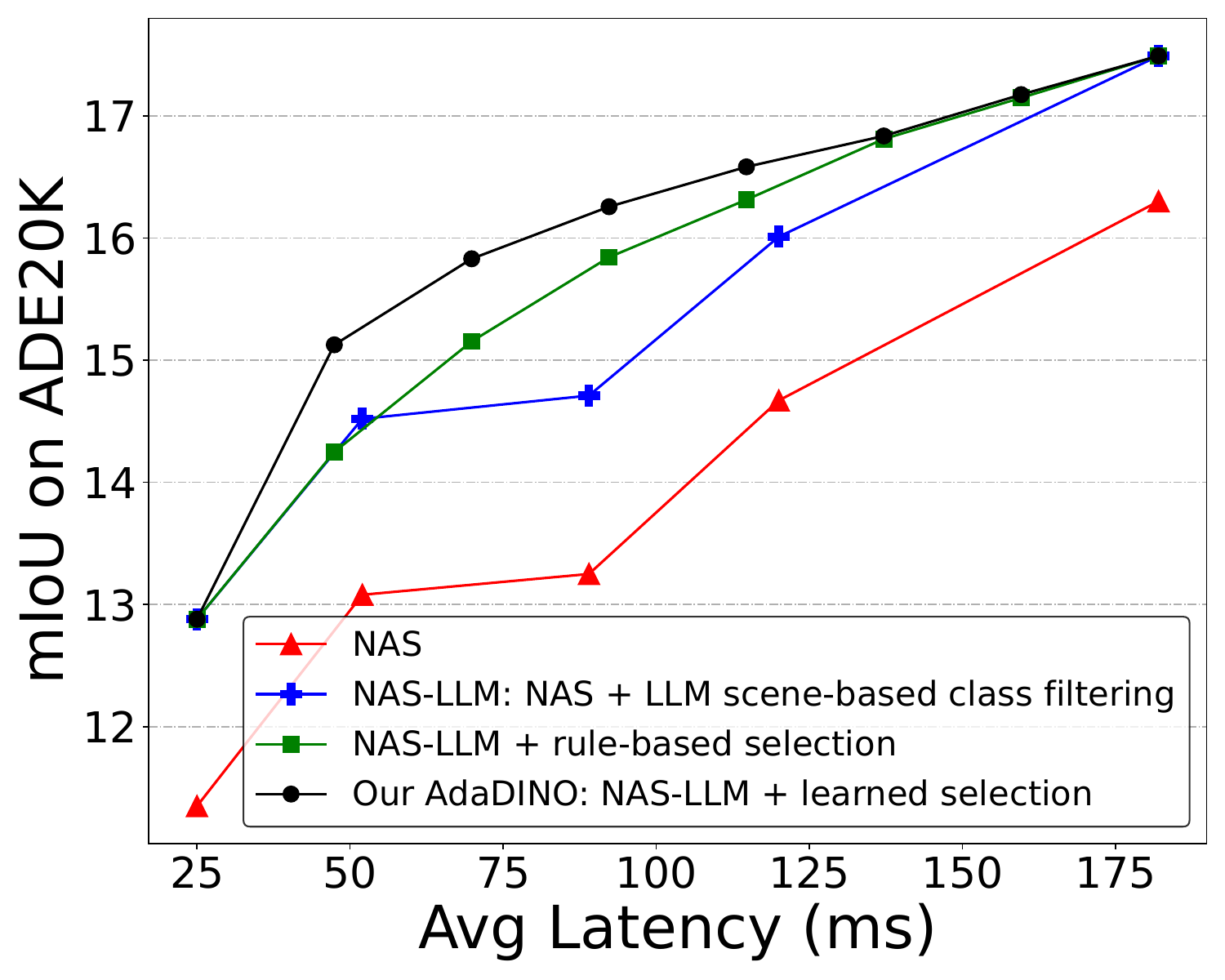}
    \vspace{-1.5em}
    \caption{Ablation of \ourname{'s} accuracy-efficiency gains, measured in average execution latency.}
    \label{fig:supp_abl_latency}
\end{figure}

\begin{figure}[t]
    \centering
    \includegraphics[width=\linewidth]{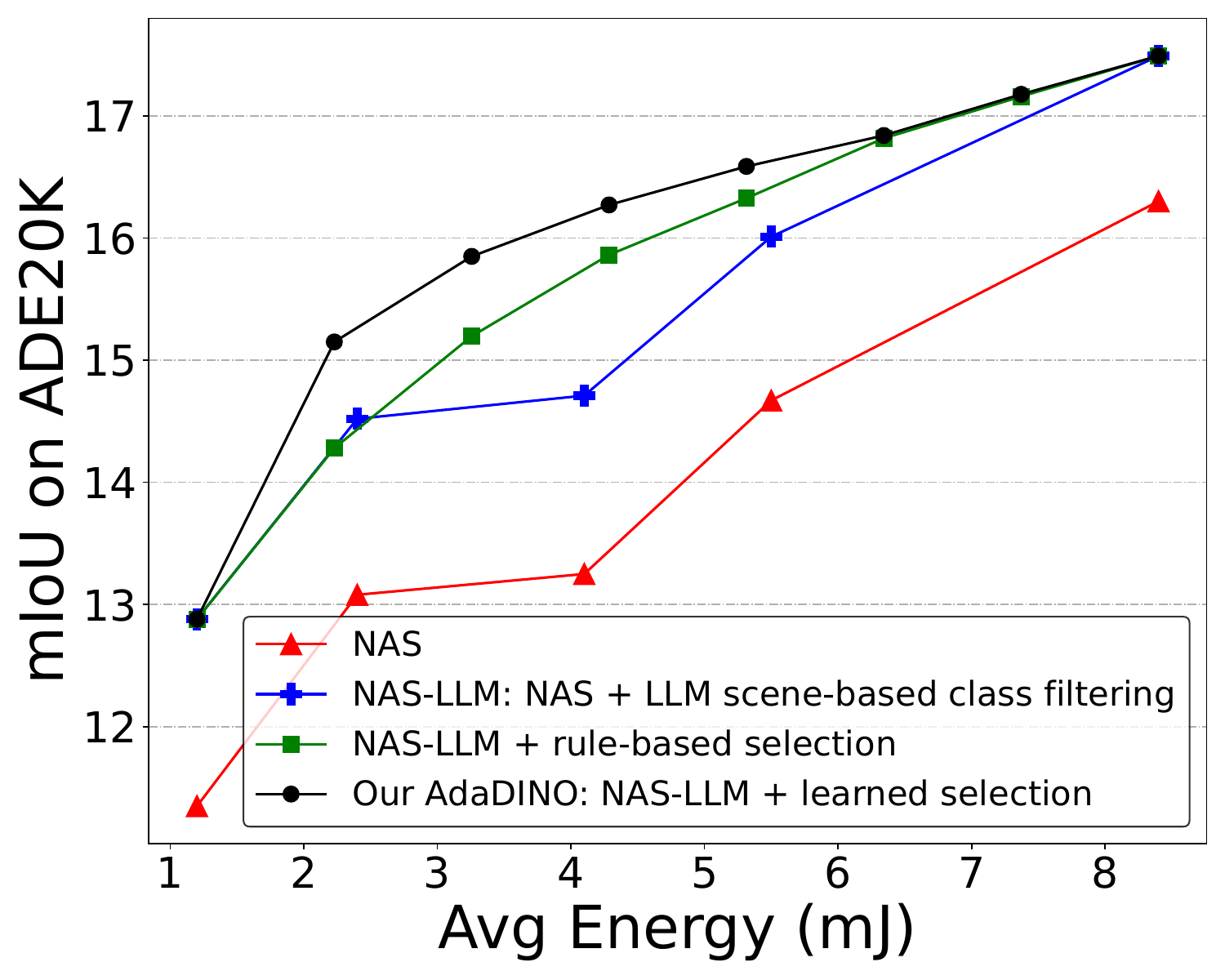}
    \vspace{-1.5em}
    \caption{Ablation of \ourname{'s} accuracy-efficiency gains, measured in average energy consumption.}
    \label{fig:supp_abl_energy}
\end{figure}

\begin{figure}[t]
    \centering
    \includegraphics[width=\linewidth]{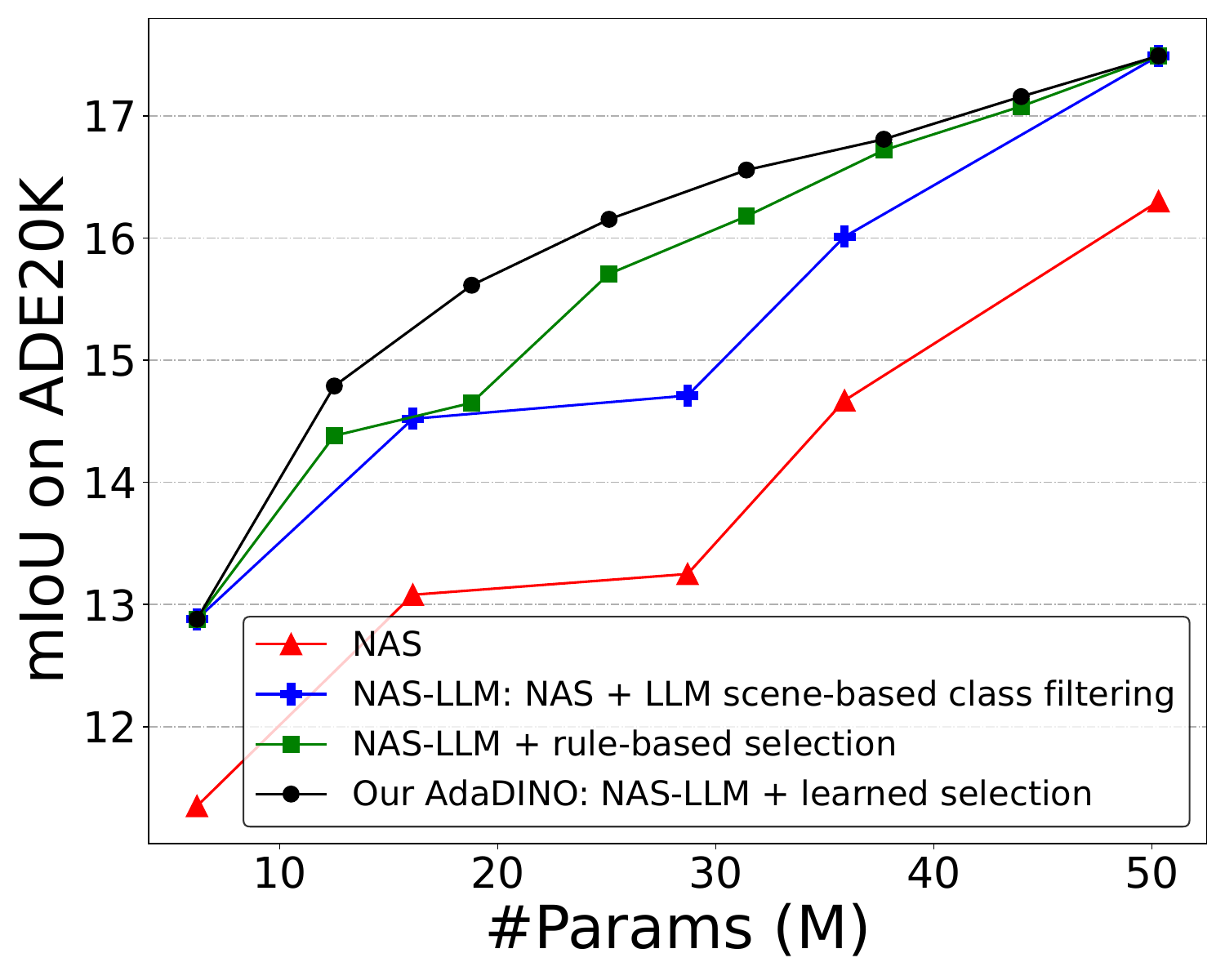}
    \vspace{-1.5em}
    \caption{Ablation of \ourname{'s} accuracy-efficiency gains, measured in number of model parameters.}
    \label{fig:supp_abl_params}
\end{figure}

\section{Extended Related Work}
\label{sec:supp_related}
This section provides an extended discussion of the related work summarized in \mainref{sec:bg}.

\subsection{Always-on Contextual AI on Edge Devices}
\label{subsec:supp_scope}
Recent advances in smartphones~\cite{georg2025fsboard}, wearable devices~\cite{abbaspourazad2024largescale}, augmented and virtual reality~\cite{abrash2021creating}, robotics~\cite{goswami2025robopepp}, and autonomous driving~\cite{lin2025hilots} have accelerated the push toward enabling AI on edge devices.
The emergence of wearable edge platforms is motivating \emph{always-on contextual AI} that can operate continuously under strict power, thermal, and memory constraints.
As illustrated in \mainref{fig:intro_motivation}, the system relies on a lightweight \emph{on-device vision foundation model} for real-time perception tasks such as classification and segmentation, due to recent progress in efficient models on mobile hardware~\cite{Mehta2022mobilevit, Liu2022convnext}.
These visual representations are aligned with language on-device to support seamless user interaction~\cite{Chen2023blip2, Li2023qformerv2}, while more computationally intensive tasks---such as multimodal reasoning, long-term memory, and access to broad world knowledge---are delegated to a cloud-based multimodal LLM~\cite{Alayrac2022flamingo, Driess2023palmE, Team2023llama2}.
This distributed edge-cloud design reflects limitations of continuous cloud streaming and aligns with prior work showing that persistent offloading is impractical for wearable devices due to energy constraints~\cite{lane2015deepx, huynh2017edgent, mittal2020survey}.
Instead, sparse visual summaries and infrequent user queries are transmitted to the cloud, enabling the construction of a personal timeline by memory-augmented multimodal systems~\cite{li2025omniquery, fan2025videoagent}.
Consequently, the always-on contextual AI requires efficient, multimodal, and text-aware foundation models capable of running on the edge while collaborating seamlessly with stronger cloud-hosted LLMs.

\myparagraph{Scope of This Work}
Bringing an always-on contextual-AI wearable to market is an undertaking at the scale of entire industry divisions, engaging hundreds to thousands of engineers and researchers.
Production efforts such as Meta Ray-Ban~\cite{meta_rayban_smart_glasses} span custom silicon and accelerator design, thermal and battery engineering, optics and display, capture and sensing stacks, interaction design, and privacy infrastructure.
Each of these is a hard engineering problem in its own right, and the open questions there---thermal envelopes, battery capacity, wearability---belong to that effort, not to any single paper.
Within this stack, the model-execution layer is a self-contained research problem, and an important one: the always-on vision workload dominates the device's sustained compute, so its efficiency directly bounds battery life and responsiveness during contextual AI usage (\mainref{subsec:method_manager}).
The layer also has well-defined inputs (the scene context and candidate class set), a well-defined output (the execution scheme of the edge model), and a measurable objective (the accuracy--efficiency frontier of \mainref{sec:eval}).
This is a scope a single research paper can treat rigorously, and precedent agrees: efficient backbones and split computing each sustain research lines of their own (\Cref{subsec:supp_adjacent}).
Concretely, given the edge--cloud division and the episodic context refresh that deployed systems adopt, we make the on-device vision model's cost track the current task, through LLM-based vocabulary refinement and learned subnet selection.
The surrounding layers---context-refresh triggering, sensing policy, and user interaction---are supplied by the deployment and compose directly with our mechanisms.
The minutes-or-longer persistence of egocentric contexts~\cite{grauman2022ego4d,grauman2024egoexo4d} makes the low-frequency refresh cadence we assume a conservative choice.

\myparagraph{Projected Deployment Benefits}
The device- and cloud-level gains this layer enables motivate the always-on setting; we report them as first-order projections under the low-frequency refresh cadence above, not as end-to-end system measurements.
Contextual-AI glasses that forward every query to the cloud~\cite{meta_rayban_smart_glasses} report query latencies around 5\,s and roughly two hours of battery life.
Running perception on the edge subnet and calling the cloud MM-LLM only when the scene changes shifts the sustained cost onto the efficient on-device model: combining our measured accelerator energy (\mainref{tab:eval_subnets}) with that cloud-only behavior projects at least $5\times$ longer battery life during contextual-AI usage, counting accelerator energy alone and excluding the camera, display, communication, and CPU.
Because the cloud is queried per context rather than per frame, cloud cost also becomes frame-rate independent---on the order of \$1.6 per hour under public API pricing against \$14 per hour-fps for cloud-only streaming, a $10$--$100\times$ reduction~\cite{openai_pricing,gemini_pricing}.
These figures are motivation for the deployment target rather than validated results, and the refresh cadence they assume is a deployment policy that our benchmark does not itself measure.

\subsection{Vision Foundation Models}
Vision foundation models have become prevalent in computer vision, offering unified representations that generalize across a wide spectrum of tasks and domains.
These models are typically trained at scale using self-supervised learning (SSL), which exploit the intrinsic structure of visual data to learn from large, unannotated image corpora.
Without explicit supervision, SSL-based foundation models can leverage virtually unlimited data sources, yielding robust and semantically rich visual embeddings.
Recent advances such as DINOv2~\cite{oquab2024dinov2}, MAE~\cite{He2022mae}, and iBOT~\cite{Zhou2022ibot} demonstrate the effectiveness of SSL in learning transferable features that rival or surpass those obtained via supervised pretraining methods like CLIP~\cite{radford2021CLIP} or PaLI~\cite{Dehghani2023pali}.
Such models have shown strong transferability, enabling a single frozen encoder to perform competitively across diverse downstream tasks without task-specific fine-tuning.
Nonetheless, large-scale vision foundation models often incur substantial computational, memory, and energy costs.
Addressing these limitations, we present \emph{adaptive vision foundation models} that achieve state-of-the-art VFM performance on edge devices.

\subsection{Contrastive Learning}
Contrastive learning methods such as CLIP enable zero-shot and open-vocabulary capabilities in VFMs by aligning visual and textual representations.
A standard CLIP architecture consists of a \VisionTower and a \TextTower, both computationally and memory intensive, making deployment on resource-constrained edge devices impractical. Pruning or shrinking them often causes significant accuracy loss~\cite{shrivastava2023clip, wu2023tinyclip}.

A key property of CLIP for edge deployment is the \defn{differing frequencies} of \VisionTower{} and \TextTower{} calls.
In always-on contextual AI (\mainref{fig:intro_motivation}), prompts or scenes change infrequently, while perception runs continuously:

\begin{itemize}
    \item \textbf{\VisionTower (high-frequency):} Processes every incoming frame and must meet strict low-latency requirements (typically $200$--$1000$~ms)~\cite{stein2021latency, xu2023ear}.
    \item \textbf{\TextTower (low-frequency):} Invoked only when user prompts or scenes change, tolerating much higher latency.
\end{itemize}

This \textit{frequency asymmetry} creates a design opportunity: the \VisionTower must be heavily optimized for real-time edge inference, while the \TextTower can remain larger without impacting responsiveness. Leveraging this imbalance is crucial for deploying CLIP-style zero-shot models efficiently on edge devices.

\myparagraph{Efficient Vision Foundation Models}
Prior work lowers VFM cost \emph{statically}, distilling or shrinking CLIP-style encoders into compact, fixed backbones~\cite{wu2023tinyclip, vasu2024mobileclip}.
\ourname{} instead keeps a family of subnets and picks among them at runtime.
Several other efficiency directions are also related to \ourname{} but adapt along a different axis.
We review them in \Cref{subsec:supp_adjacent} and clarify how their scope differs from ours.

\subsection{Relation to Adjacent Directions}
\label{subsec:supp_adjacent}
Several established research directions are close to \ourname{} in spirit but differ in scope.
We summarize the most relevant ones below and, for each, explain why it is either outside the problem \ourname{} addresses or orthogonal to our design and composable with it.
In every case the two are complementary rather than competing, so these methods are not substitutes that our evaluation would need to compare against.

\myparagraph{Adaptive and Dynamic Inference}
A large body of work makes a single network input-adaptive, spending more or less computation on each input according to its estimated difficulty~\cite{han2021dynamic}.
Typical strategies include token pruning, merging, and sparsification in vision transformers~\cite{liang2022evit, bolya2023tome, rao2021dynamicvit, wu2025patchranking, wang2021notall, chen2023cfvit}, token or block halting~\cite{yin2022avit, meng2022adavit}, input-dependent depth and layer skipping~\cite{wang2018skipnet, wu2018blockdrop}, early-exit and anytime prediction~\cite{huang2018msdnet, teerapittayanon2016branchynet}, runtime width selection within one supernet~\cite{yu2019slimmable, li2021dsnet}, resolution-adaptive computation~\cite{yang2020ranet, figurnov2017sact, wang2020glance}, sparsely-gated mixtures of experts~\cite{shazeer2017moe, fedus2022switch, riquelme2021vmoe}, adaptive frame selection for video~\cite{wu2019adaframe}, and budget-aware or token-reduced inference for multimodal LLMs~\cite{xu2025adallava, chen2024fastv, zhang2025sparsevlm, xing2025pyramiddrop, hu2024mqtllava}.
These methods read a per-input signal, such as prediction confidence or token saliency, and adjust how much of one fixed, closed-vocabulary model to run.
\ourname{} instead chooses among distinct open-vocabulary subnets from the scene and the user-defined class vocabulary, which describe the task rather than an individual frame.
The two are independent, and any of these per-input mechanisms can still run inside the subnet \ourname{} activates, so they complement rather than replace our adaptive selection.

\myparagraph{Model Compression and Quantization}
Pruning, quantization, and knowledge distillation~\cite{han2016deepcompression, jacob2018quantization, hinton2015distilling, touvron2021deit, yuan2022ptq4vit, yang2024clipkd} shrink a network to a single cheaper operating point.
Our subnets are themselves distilled from a large teacher, and each one can be pruned or quantized on its own.
Compression produces one smaller model, whereas \ourname{} maintains several and decides which to run for a given task.

\myparagraph{Collaborative Edge-Cloud Inference}
Split computing divides a single network between the device and the cloud, or exits early on-device, to balance latency, energy, and accuracy~\cite{kang2017neurosurgeon, laskaridis2020spinn, matsubara2022split, banitalebidehkordi2021autosplit, zeng2020coedge}.
\ourname{} does not split the vision model: the selected subnet runs entirely on the edge, and only the infrequent decision of scene understanding and vocabulary filtering is sent to the cloud LLM.
The design question is thus which computation to offload and how often, not where to cut a fixed network.

\myparagraph{Cascades and Input-Dependent Routing}
Cascades and routers escalate or dispatch each input among several separate models based on confidence or cost, in both vision and language settings~\cite{bolukbasi2017adaptive, jitkrittum2023when, chen2023frugalgpt, ong2024routellm, ding2024hybrid}.
\ourname{} selects among subnets of one weight-sharing supernet rather than a pool of independent models, and routes on scene and task semantics for open-vocabulary perception rather than on the confidence of a fixed label set.
It also decides once per scene, so no input is re-run through a larger model.

\myparagraph{LLM-Driven Vision Systems}
Several systems use an LLM as a controller that plans and calls vision tools for each query~\cite{shen2023hugginggpt, suris2023vipergpt, wu2023visualchatgpt, gupta2023visual, lu2023chameleon}.
In \ourname{} the LLM has a narrower, low-frequency role: it summarizes the scene and filters the class vocabulary for the on-device model, and stays out of the per-frame perception loop.
The contribution here is the edge vision model and how it is chosen, not an agent that reasons about every input.

\myparagraph{Open-Vocabulary Segmentation}
Specialized open-vocabulary segmentation and detection models add dedicated decoders or adapters on top of large backbones to push accuracy~\cite{li2022lseg, ghiasi2022scaling, xu2023san, cho2024catseg, xu2023odise, liang2023ovseg, yu2023fcclip, gu2022vild}.
They aim for segmentation quality at server-scale compute, whereas \ourname{} uses open-vocabulary segmentation as one way to probe an edge backbone under a tight compute budget, following the protocol of DINO.txt~\cite{jose2025dinov2}.
We report against that open-vocabulary VFM, which shares our setting.
The heavier, task-specialized models sit in a different accuracy-compute regime.

\myparagraph{Prompt and Vocabulary Design for Open-Vocabulary Models}
A related line improves the text side of open-vocabulary recognition, for instance by enriching class names with LLM-generated descriptions~\cite{pratt2023does, menon2023visual}, ensembling or reweighting prompts~\cite{roth2023waffling}, or learning prompts and adapters~\cite{mirza2023lafter, khattak2025learning, zhou2022learning, zhou2022conditional, wang2024learning}.
These act on a fixed target vocabulary and are computed offline, whereas the LLM in \ourname{} selects and filters the active vocabulary at runtime from the inferred scene.
The two are complementary, and several such methods already appear as baselines in our accuracy-efficiency comparison~\cite{xing2023rewrite, xu2023learning, roth2023waffling, mirza2023lafter, khattak2025learning}.

\myparagraph{Summary}
Across these directions, each is either outside the problem \ourname{} targets or a complementary optimization that composes with our subnets, so none is a substitute for the task-level, open-vocabulary model selection we introduce.
The comparisons that isolate this axis---adaptive selection versus fixed backbones, and our learned LLM-guided selector versus rule-based and non-LLM alternatives---are already reported in \Cref{sec:supp_rule_selector}, \Cref{subsec:supp_ood_scenes}, and \mainref{fig:abl_manager}.
Our evaluation therefore covers the comparisons relevant to our claims.

\subsection{Neural Architecture Search}
Neural Architecture Search (NAS) provides a principled framework for adaptive-capacity models, suited to resource-constrained edge deployment.
Its key merit is support for \textbf{\textit{runtime subnet selection}} over sub-architectures of varying computational cost.
At inference, it selects a subnet by task complexity or resources, enabling flexible accuracy-efficiency trade-offs.
While early NAS methods used evolutionary~\cite{real2017large} or RL-based strategies~\cite{zoph2016neural}, later weight-sharing supernets~\cite{cai2020once,chu2021fairnas,yu2020bignas} made NAS practical for adaptiveness.
Slimmable networks select widths at run time under resource budgets~\cite{yu2019slimmable,li2021dsnet}, an early form of multi-capacity runtime switching.
These supernets usually pick a subnet to meet a hardware or latency budget.
In \ourname{}, the choice is instead driven by task semantics: the LLM manager reads the scene and class vocabulary, and its learned selector picks the subnet.
We use NAS to develop text-aligned, adaptive VFMs for always-on edge scenarios.

\else\ifsplitparts
\maketitle
\begin{bibunit}
\begin{abstract}
Always-on contextual AI runs language-aligned vision foundation models (VFMs) on edge devices, where the on-device model is the dominant continuous compute cost under strict latency and power limits.
Due to an observed low-frequency shift in scene context and its relevant vocabulary, we present \ourname, an adaptive framework that makes on-device VFM inference efficient by matching execution to the current scene and task.
We build on a known phenomenon, that the accuracy drop of shrinking model sizes depends on the task, and turn it into task-level adaptive execution.
\ourname integrates neural architecture search (NAS) into a language-aligned VFM backbone distilled from DINOv2, training a single family of subnets for efficient execution during runtime.
A multimodal large language model (LLM) on the cloud, invoked at low frequency, refines the candidate class set from scene context, while a learned selector activates the least-cost subnet predicted to retain a target fraction of accuracy.
With the backbone and semantic pipeline held fixed, learned selection alone reduces average compute by $37\%$ over the best fixed subnet at equal segmentation accuracy.
Across zero-shot classification and open-vocabulary segmentation, \ourname establishes a strong accuracy-efficiency frontier, improving over evaluated models of comparable sizes by up to $7.9\%$ in acc@1 on IN1K and $5.2\%$ mIoU on ADE20K, and reducing average FLOPs by up to $74.9\%$ at similar accuracy.
\end{abstract}

\section{Conclusion}
\label{sec:concl}
We present \ourname{}, a context-adaptive framework for efficient edge DINOv2-distilled VFMs that combines a NAS-enabled model family with an LLM-assisted runtime manager to refine the active vocabulary and select the least-cost subnet by task context.
With backbone and semantic pipeline held fixed, learned selection alone cuts average compute by 37\%, and overall it improves the accuracy-efficiency trade-off by up to 5.2\% mIoU or 74.9\% fewer FLOPs---task-level adaptation built on standard, well-optimized components that stay deployable on edge accelerators.
Next steps include context-refresh scheduling and alternative context providers.

\FloatBarrier
{\small %
\putbib
}
\end{bibunit}
\clearpage
\setcounter{figure}{0}\setcounter{table}{0}%
\setcounter{equation}{0}\setcounter{footnote}{0}%
\twocolumn[%
  \vbox to \titlebox{%
    \hsize\textwidth\linewidth\hsize%
    \vskip 0.625in minus 0.125in%
    \centering%
    {\LARGE\bf \realTitle\xspace(Supplementary Material)\par}%
    \vskip 0.1in plus 0.5fil minus 0.05in%
    {\Large\textbf{Anonymous submission}}%
    \vskip 1em plus 2fil%
  }%
]%
\begin{bibunit}

\FloatBarrier
{\small %
\putbib
}
\end{bibunit}

\else
\maketitle
\ifshowmain
\begin{abstract}
Always-on contextual AI runs language-aligned vision foundation models (VFMs) on edge devices, where the on-device model is the dominant continuous compute cost under strict latency and power limits.
Due to an observed low-frequency shift in scene context and its relevant vocabulary, we present \ourname, an adaptive framework that makes on-device VFM inference efficient by matching execution to the current scene and task.
We build on a known phenomenon, that the accuracy drop of shrinking model sizes depends on the task, and turn it into task-level adaptive execution.
\ourname integrates neural architecture search (NAS) into a language-aligned VFM backbone distilled from DINOv2, training a single family of subnets for efficient execution during runtime.
A multimodal large language model (LLM) on the cloud, invoked at low frequency, refines the candidate class set from scene context, while a learned selector activates the least-cost subnet predicted to retain a target fraction of accuracy.
With the backbone and semantic pipeline held fixed, learned selection alone reduces average compute by $37\%$ over the best fixed subnet at equal segmentation accuracy.
Across zero-shot classification and open-vocabulary segmentation, \ourname establishes a strong accuracy-efficiency frontier, improving over evaluated models of comparable sizes by up to $7.9\%$ in acc@1 on IN1K and $5.2\%$ mIoU on ADE20K, and reducing average FLOPs by up to $74.9\%$ at similar accuracy.
\end{abstract}

\section{Conclusion}
\label{sec:concl}
We present \ourname{}, a context-adaptive framework for efficient edge DINOv2-distilled VFMs that combines a NAS-enabled model family with an LLM-assisted runtime manager to refine the active vocabulary and select the least-cost subnet by task context.
With backbone and semantic pipeline held fixed, learned selection alone cuts average compute by 37\%, and overall it improves the accuracy-efficiency trade-off by up to 5.2\% mIoU or 74.9\% fewer FLOPs---task-level adaptation built on standard, well-optimized components that stay deployable on edge accelerators.
Next steps include context-refresh scheduling and alternative context providers.

\fi
\ifshowsupp

\fi
{\small
\bibliography{ref}
}

\fi\fi

\end{document}